\documentclass[Afour,sageh,times]{sagej}

\usepackage{amsmath,amssymb,mathrsfs,amsfonts,bm}
\usepackage[ruled,vlined]{algorithm2e}
\usepackage{graphicx}
\usepackage{epstopdf}
\usepackage{subcaption}
\usepackage{color}
\usepackage{multirow}
\usepackage{siunitx}
\usepackage{rotating}

\newtheorem{remark}{Remark}

\newcommand{\Ad}{\mathrm{Ad}}
\newcommand{\ad}{\mathrm{ad}}

\newcommand{\M}{M}
\newcommand{\K}{K}
\newcommand{\F}{F}
\newcommand{\cross}{^\wedge}
\newcommand{\ve}{^\vee}

\newcommand{\diag}{\mathrm{diag}}

\begin{document}

\title{Estimating Dynamic Soft Continuum Robot States from Boundaries}

\author{Tongjia Zheng and Jessica Burgner-Kahrs}

\affiliation{The authors are with the Continuum Robotics Laboratory and the Robotics Institute at the University of Toronto, ON, Canada. 
}

\corrauth{Tongjia Zheng}

\email{tongjia.zheng@utoronto.ca}

\begin{abstract}
State estimation is one of the fundamental problems in robotics. For soft continuum robots, this task is particularly challenging because their states (poses, strains, internal wrenches, and velocities) are inherently \textit{infinite-dimensional} due to continuous deformability, while sensing provides only discrete measurements. Recently, a dynamic state estimation method known as a \textit{boundary observer} was introduced, which uses Cosserat rod theory to recover all states from tip velocity measurements. In this work, we present a dual design that instead relies on measuring the internal wrench at the robot's base. Despite the duality, this approach offers a key practical advantage: it requires only a force/torque (FT) sensor at the base and eliminates the need for external motion capture systems. Both observer types are inspired by energy dissipation principles and can be combined to enhance performance. We conduct a Lyapunov-based analysis to study convergence and reveal a useful property: as observer gains increase, the convergence rate first improves and then degrades. This convex trend enables efficient gain tuning. We also identify cases where linear and angular states are fully determined by each other, further relaxing sensing requirements. In summary, this work achieves dynamic infinite-dimensional state estimation with minimal sensing requirements and systematic parameter tuning, which has not been demonstrated in existing approaches. Simulation and experimental studies using a tendon-driven continuum robot validate convergence under fast dynamic motions, the existence of optimal gains, robustness to external forces, measurement noise, and model uncertainty, and real-time computational performance.
\end{abstract}


\maketitle

\section{Introduction}

Soft continuum robots, inspired by the movements of snakes, worms, and elephant trunks \citep{robinson1999ContinuumRobotsStatea}, feature a slender, flexible design that enables them to navigate complex, confined spaces where traditional rigid robots struggle to operate. This adaptability has opened up new possibilities for precise interventions in challenging environments, including medical surgeries \citep{burgner2015continuum} and inspection or repair tasks in constrained settings \citep{russo2023ContinuumRobotsOverview}.

For many robotic applications, accurately knowing the robot's current state is essential. However, state estimation for soft continuum robots is particularly challenging. Unlike rigid robots, soft continuum robots can deform continuously, and their state can be conceptualized as a continuous set of rigid cross-sections along a centerline. This means that the state includes not only positions, orientations, linear velocities, and angular velocities but also parameters such as bending, torsion, shear, and elongation. These state variables are continuous functions of arc length and time, making the problem inherently infinite-dimensional and thus more complex than for traditional rigid robots.

Since traditional sensing techniques provide only discrete measurements, one common strategy for robot state estimation is to combine these discrete measurements with mathematical models to infer the continuous states that are not directly measurable. Most existing research has focused on shape estimation, which assumes quasi-static conditions to estimate the configuration of the entire soft continuum robot. The complexity of these models varies, from parameterized spatial curve fitting \citep{song2015electromagnetic, bezawada2022shape} to fitting Cosserat rod statics \citep{sadati2020stiffness, feliu2025actuation}, Kalman filtering based on Kirchhoff rod statics \citep{anderson2017continuum}, and Gaussian process regression for Cosserat rod kinematics \citep{lilge2022continuum, ferguson2024unified}. In particular, the approaches based on rod models leverage the analogy between trajectory estimation for mobile robots over time and shape estimation for soft continuum robots along arclength. This similarity enables the adaptation of many successful state estimation algorithms from mobile robots to continuum robots.

\begin{table*}[]
\small
\setlength{\tabcolsep}{0.35em}
\renewcommand{\arraystretch}{1.5}
    \centering
    \caption{Overview of related work on continuum robot state estimation.}
    \label{tab:survey}
    \begin{tabular}{llp{9em}p{9em}p{5.5em}p{13em}}
        \hline
        Reference & Type & Model & Measurements & Algorithm & Estimated States \\
        \hline
        \citep{song2015electromagnetic} & Static & Cubic Bézier curve & Pose at multiple points & Curve fitting & Pose along arclength \\
        \citep{bezawada2022shape} & Static & Pythagorean Hodograph curves & Position/orientation at multiple points & Curve fitting & Pose along arclength \\
        \citep{anderson2017continuum} & Static & Kirchhoff rod statics & Pose at multiple points & Kalman filter & Pose along arclength, external force \\
        \citep{sadati2020stiffness} & Static & Cosserat rod statics & Base wrench, chamber pressure & ODE solver & Pose along arclength, tip wrench \\
        \citep{lilge2022continuum} & Static & Cosserat rod kinematics & Pose and strain at multiple points & Gaussian process regression & Pose and strain along arclength \\
        \citep{ferguson2024unified} & Static & Cosserat rod kinematics & Pose and strain at multiple points & Gaussian process regression & Pose, strain, and external wrench along arclength \\
        \citep{feliu2025actuation} & Static & Reduced-order Cosserat rod statics & Tendon tension and displacement & Nonlinear optimization & Pose and external force along arclength \\
        \hline
        \citep{loo2019h} & Dynamic & Constant curvature & Tip position, chamber pressure & Extended Kalman filter & Constant curvature arc parameter along time \\
        \citep{rucker2022task} & Dynamic & Discrete planar Kirchhoff rod dynamics & Tendon tension, tip position and velocity & Observer & Discrete pose, velocity, and internal wrench along arclength and time \\
        \citep{feliu2024dynamic} & Dynamic & Reduced-order Cosserat rod dynamics & Tendon tension and length & State-dependent Kalman filter & Pose, velocity, and internal wrench along arclength and time \\
        \citep{zheng2024full} & Dynamic & Cosserat rod dynamics & Tendon tension, tip velocity & Boundary observer & Pose, velocity, and internal wrench along arclength and time \\
        \citep{zheng2024estimating} & Dynamic & Cosserat rod dynamics & Tendon tension, tip pose and velocity & Boundary observer & Pose, velocity, and internal wrench along arclength and time \\
        This work & Dynamic & Cosserat rod dynamics & Tendon tension, base wrench & Boundary observer & Pose, velocity, and internal wrench along arclength and time \\
        \hline
    \end{tabular}
\end{table*}


Shape estimation, however, is limited to slow-speed motions, and performance degradation has been observed in dynamic tests \citep{feliu2025actuation}. Consequently, there has been a growing interest in incorporating dynamic models. Among these, Cosserat rod theory is perhaps the most widely adopted model for soft continuum robots \citep{simo1988dynamics, rucker2011statics, renda2014dynamic}. However, this approach leads to nonlinear partial differential equations (PDEs) evolving in the geometric Lie group $SE(3)$, making state estimation particularly challenging. To address this, researchers have developed various discretized or reduced-order models, such as finite difference \citep{rucker2022task}, piecewise constant curvature \citep{della2020model}, piecewise constant strain \citep{renda2018discrete}, and smooth strain parameterizations \citep{boyer2020dynamics}. Building on these discretized or reduced-order models, techniques like extended Kalman filters \citep{loo2019h}, state-dependent Kalman filters \citep{feliu2024dynamic}, and passivity-based observers \citep{rucker2022task} have been designed. More recently, a dynamic state estimation algorithm based directly on the original Cosserat rod PDEs, without any initial discretization or order reduction, was introduced in~\citep{zheng2024full, zheng2024estimating}. This method recovers the full set of infinite-dimensional dynamic states using only tip velocity (and pose) measurements. We refer to this method as a \textit{tip observer}.


This work is motivated by the need to advance infinite-dimensional dynamic state estimation through alternative embedded sensing modalities, to further reduce sensing requirements in specific scenarios, and to provide insight into the impact of algorithmic parameters for systematic tuning. In particular, we present a novel boundary observer that can also recover infinite-dimensional dynamic states, but instead relies on measuring the internal wrench (force and moment) at the robot's base. We refer to this algorithm as a \textit{base observer}. It compares the estimated wrench with the measured value at the base and injects the discrepancy as a virtual velocity input, thereby ``swinging'' the observer in a way that dissipates estimation error energy. Despite the mathematical symmetry, this approach offers a practical advantage: it requires only a force/torque sensor embedded at the base, eliminating the need for external motion capture systems. Combining both base and tip boundary corrections also yields a generalized form of the boundary observer.
We then perform a Lyapunov-based analysis to study the convergence of these observers. A key finding is a convex relationship between observer gains and convergence rate: increasing the gains initially accelerates convergence, but beyond a certain point, the rate deteriorates. This convex behavior facilitates efficient gain tuning. We also show that in special cases, the linear and angular states are fully determined by each other, thus further relaxing sensing requirements.
Simulation and experimental studies on a tendon-driven continuum robot verify the effectiveness of all observer variants in tracking fast dynamic motions, the ``increase-then-decrease'' trend in convergence time with respect to gain tuning, and robustness against unknown external forces, measurement noise, and model uncertainty.

\medskip
\noindent\textbf{Contributions.} Table~\ref{tab:survey} presents a comparison with representative related work based on factors most relevant to state estimation, including whether the formulation is static or dynamic, the underlying model assumptions, sensing requirements, and the states that can be reconstructed. This work is distinguished from traditional static shape estimation methods by its fully dynamic formulation. Although the use of a base force sensor has previously been explored in a static context~\citep{sadati2020stiffness}, our method addresses the dynamic case. Compared with the dynamic estimation algorithms in~\citep{rucker2022task, feliu2024dynamic}, the key novelty lies in the fact that our observer is derived directly from the full Cosserat rod PDEs, without initial discretization or order reduction. This distinction is important because performance degradation has been observed in reduced-order estimation algorithms as the model order increases~\citep{feliu2024dynamic}.
This work is most closely related to our prior work~\citep{zheng2024estimating, zheng2024full}, but extends it in three directions: (i) introducing a novel embedded-sensor-based design and yielding a more general boundary observer, (ii) identifying state redundancy which can be exploited to further relax sensing requirements, and (iii) conducting a deeper theoretical analysis of optimal observer design and fastest convergence rate.
In summary, this work achieves dynamic full-state estimation using the complete nonlinear model with minimal sensing requirements and systematic parameter tuning, thereby distinguishing itself from existing approaches in at least one of these aspects.

Lastly, it is worth pointing out that while some static estimation algorithms can estimate external forces via optimization, this has not been achieved in dynamic settings, where external forces are typically time-varying. Extending observer frameworks to dynamically estimate such forces remains an important direction for future research.

\section{Formulation of the State Estimation Problem}
\label{sec:modeling}

Our objective is to estimate the infinite-dimensional dynamic states of a tendon-driven continuum robot.
Although we use tendon-driven continuum robots as an example, the algorithm can potentially be extended to other robots, such as soft continuum manipulators, and to other actuation mechanisms, such as fluidic actuators.
Our algorithm is grounded in the Cosserat rod theory \citep{rucker2011statics, renda2014dynamic}, a widely adopted dynamic model for soft continuum robots. We employ Lie group notation in $SO(3)$ and $SE(3)$ \citep{murray2017mathematical}. Detailed definitions are provided in the Appendix.

\begin{figure}[t]
    \centering
    \includegraphics[width=0.9\columnwidth]{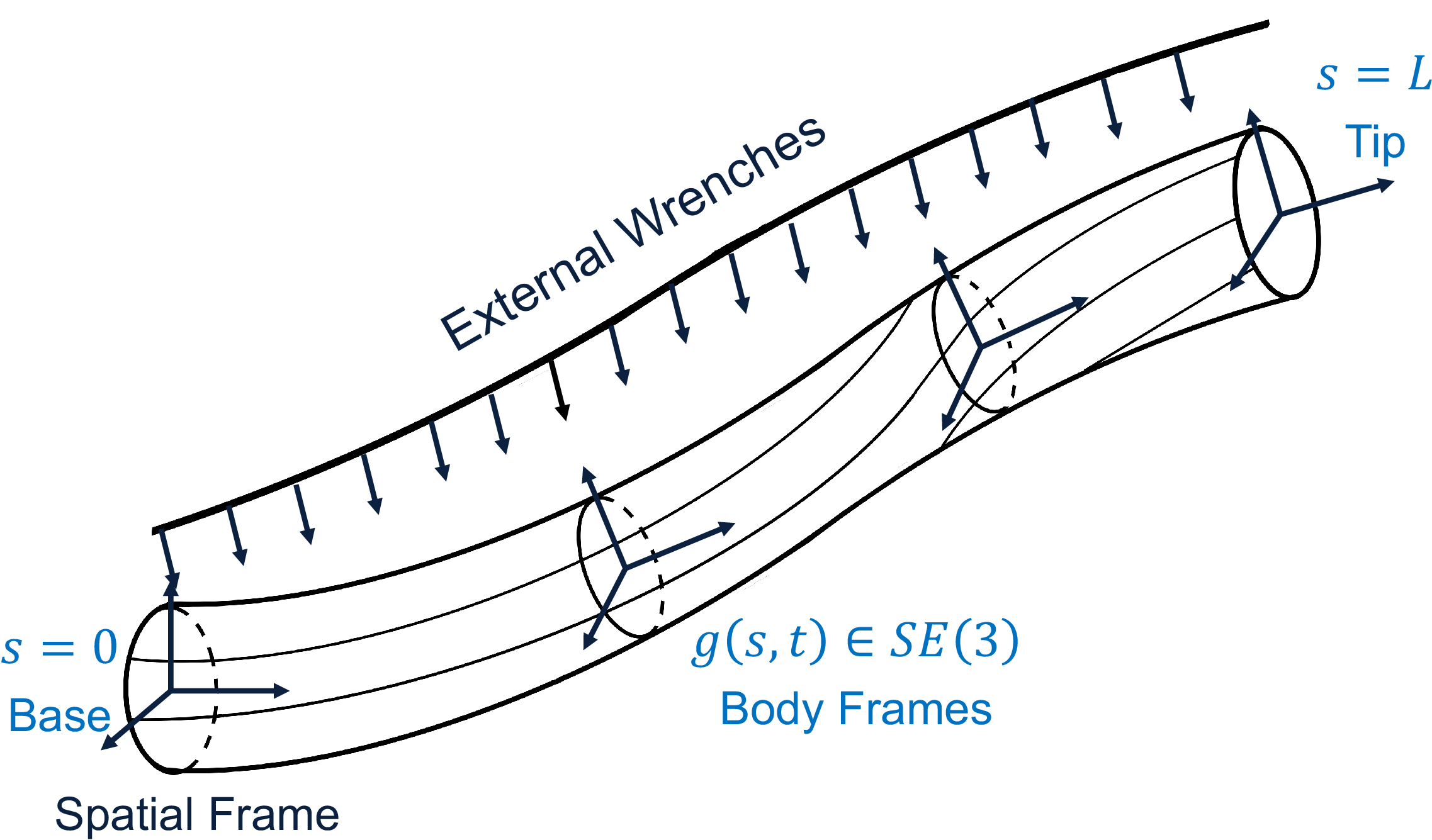}
    \caption{Using Cosserat rod theory, the robot is modeled as a continuous set of rigid cross-sections stacked along a centerline parametrized by $s$ from the base ($s=0$) to the tip ($s=L$). $g(s,t)\in SE(3)$ represents the pose of the cross-section at location $s$ and time $t$.
    }
    \label{fig:Cosserat rod}
\end{figure}

\begin{table}[]
\small
\setlength{\tabcolsep}{2pt}
\renewcommand{\arraystretch}{1.2}
    \centering
    \caption{Nomenclature in the Cosserat rod model}
    \label{tab:nomenclature}
    \begin{tabular}{rl}
        \hline
        $s\in[0,L]$ & arc parameter \\
        $t\in\mathbb{R}$ & time \\
        $(\cdot)_s,~(\cdot)_t$ & partial derivatives \\
        $p(s,t)\in\mathbb{R}^3$ & position \\
        $R(s,t)\in SO(3)$ & rotation \\
        $g = (R,p)\in SE(3)$ & pose \\
        $v(s,t)\in\mathbb{R}^3$ & linear velocity \\
        $w(s,t)\in\mathbb{R}^3$ & angular velocity \\
        $\eta = [w~v]^\top$ & velocity twist \\
        $q(s,t),q_o(s)\in\mathbb{R}^3$ & linear strain, reference linear strain \\
        $u(s,t),u_o(s)\in\mathbb{R}^3$ & angular strain, reference angular strain \\
        $\xi = [u~q]^\top, \xi_o = [u_o~q_o]^\top$ & strain twist, reference strain twist \\
        $n(s,t)\in\mathbb{R}^3$ & internal force \\
        $m(s,t)\in\mathbb{R}^3$ & internal moment \\
        $\Lambda = [m~n]^\top$ & total internal wrench \\
        $\Lambda_{\text{elast}}(s,t)$ & elastic wrench \\
        $\Lambda_{\text{act}}(s,t)$ & applied internal wrench by actuators \\
        $f(s,t)\in\mathbb{R}^3$ & external force \\
        $l(s,t)\in\mathbb{R}^3$ & external moment \\
        $\F = [l~f]^\top$ & external wrench \\
        $g_0(t)$, $\eta_0(t)$ & applied base pose and velocity \\
        $\F_1(t)$ & applied tip wrench \\
        $\bar{\Lambda}_0(t)$ & base wrench measurement \\
        $\bar{g}_1(t)$, $\bar{\eta}_1(t)$ & tip pose and velocity measurement \\
        $\Gamma_0$, $\Gamma_1$, $\Gamma_{\text{P}}$, $\Gamma_{\text{D}}$ & observer gains \\
        $\M_l(s)\in\mathbb{R}^{3\times3}$ & linear inertia matrix \\
        $\M_a(s)\in\mathbb{R}^{3\times3}$ & angular inertia matrix \\
        $\M=\diag(\M_a,\M_l)$ & inertia matrix \\
        $K_l(s)\in\mathbb{R}^{3\times3}$ & linear stiffness matrix \\
        $K_a(s)\in\mathbb{R}^{3\times3}$ & angular stiffness matrix \\
        $\K=\diag(\K_a,\K_l)$ & stiffness matrix \\
        \hline
    \end{tabular}
\end{table}

Cosserat rod theory models a soft continuum robot as a continuous set of rigid cross-sections stacked along a centerline.
An illustration is given in Fig.~\ref{fig:Cosserat rod}. The nomenclature is provided in Table~\ref{tab:nomenclature}.
The states of the robot are continuous functions of the arc parameter $s\in[0,L]$ and time $t$, where $L$ is the total length.
Let $p(s,t)\in\mathbb{R}^3$ be the position vector of the centerline and $R(s,t)\in SO(3)$ be the rotation matrix of each cross-section.
Each cross-section thus defines a body frame.
The pose of the robot is denoted by
\begin{align}
    g(s,t)=
    \begin{bmatrix}
        R(s,t) & p(s,t) \\
        0_{1\times3} & 1
    \end{bmatrix}\in SE(3),
\end{align}

To define kinematics, let $w(s,t),v(s,t),u(s,t),q(s,t)\in\mathbb{R}^3$ be the fields of angular velocities, linear velocities, angular strains, and linear strains, respectively, of the cross-sections in their body frames.
Let
\begin{align}
    \eta(s,t)=
    \begin{bmatrix}
        w(s,t) \\
        v(s,t)
    \end{bmatrix},\quad
    \xi(s,t)=
    \begin{bmatrix}
        u(s,t) \\
        q(s,t)
    \end{bmatrix}
\end{align}
be the fields of velocity twists and strain twists, respectively.
We use $(\cdot)_t$ and $(\cdot)_s$ to denote partial derivatives and omit the dependence on $s$ and $t$ for simplicity.
The kinematics of the soft continuum robot are given by
\begin{align}
    g_t=g\eta^\wedge, \label{eq:kinematics} \\
    g_s=g\xi^\wedge, \label{eq:forward kinematics}
\end{align}
The equality of mixed partial derivatives $\partial_{st}g=\partial_{ts}g$ yields the compatibility equation:
\begin{align}
    \xi_t=\eta_s+\ad_{\xi}\eta. \label{eq:compatibility}
\end{align}

To define dynamics, let $m(s,t),n(s,t),l(s,t),f(s,t)\in\mathbb{R}^3$ be the fields of internal moments, internal forces, external moments, and external forces, respectively, of the cross-sections in their body frames.
Let
\begin{align}
    \Lambda(s,t)=
    \begin{bmatrix}
        m(s,t) \\
        n(s,t)
    \end{bmatrix},\quad
    \F(s,t)=
    \begin{bmatrix}
        l(s,t) \\
        f(s,t)
    \end{bmatrix}
\end{align}
be the fields of internal and external wrenches, respectively.
Using Hamilton's principle in the context of Lie groups, the robot dynamics (expressed in body frames) are given by
\begin{align} \label{eq:dynamics}
    \M\eta_t-\ad_\eta^\top\M\eta=\Lambda_s-\ad_\xi^\top\Lambda+\F,
\end{align}
where $\M(s)\in\mathbb{R}^{6\times6}$ is the cross-sectional inertia matrix, which may vary along $s$, for example, in cases of non-uniform cross-sectional areas.
Since, in practice, $\F$ primarily represents gravitational effects, which are defined in the global frame, it is expressed as
\begin{align} \label{eq:external wrench}
    \F(s,t) = \Ad_{g(s,t)}^{-1}\F_G(s),
\end{align}
where $\F_G(s)$ denotes the wrench field due to gravity.

Tendon actuation is modeled as part of the distributed internal wrenches \citep{renda2020geometric}, resulting in
\begin{align} \label{eq:total internal wrench}
    \Lambda(s,t) & =\Lambda_{\text{elast}}(s,t)+\Lambda_{\text{act}}(s,t),
\end{align}
where $\Lambda_{\text{elast}}(s,t)$ represents the internal wrench due to elastic deformation and $\Lambda_{\text{act}}(s,t)$ denotes the internal wrench generated by tendon actuation.
The wrench $\Lambda_{\text{elast}}$ depends on the strain $\xi$ through a constitutive law, such as Hooke's law:
\begin{align} \label{eq:linear constitutive law}
    \Lambda_{\text{elast}}=\K(\xi-\xi_o),
\end{align}
where $\xi_o(s)\in\mathbb{R}^6$ is the strain field of the undeformed reference configuration and $\K(s)\in\mathbb{R}^{6\times6}$ is the cross-sectional stiffness matrix, which may vary along $s$.
Here $\xi_o(s)\equiv[0~0~0~0~0~1]^\top$ if the reference configuration is straight.
More general constitutive laws may be used to incorporate viscosity and nonlinear stress-strain relations.

The actuation wrench $\Lambda_{\text{act}}(s,t)$ can be determined using an actuator model as described in \citep{renda2020geometric}. For tendon-driven soft continuum robots, this model can be simplified, as outlined in \citep{boyer2022statics}.
Let $d_i(s)\in\mathbb{R}^3$ be the position at which tendon $i$ intersects the cross-section at $s$, expressed in the body frame, and let $d_i'(s)$ be its spatial derivative.
Let $\tau_i(t)$ be the tendon tension, which is always positive.
Assuming that the tendons are frictionless,
\begin{align}\label{eq:actuator model}
    \Lambda_{\text{act}}(s,t)=\sum_{i=1}^2\frac{\tau_i(t)}{\|T_i(s)\|}
    \begin{bmatrix}
        d_i(s)\cross T_i(s) \\
        T_i(s)
    \end{bmatrix}
\end{align}
where $T_i/\|T_i\|$ is the unit tangent vector of the tendon expressed in the body frames with $T_i=q_o+u_o\cross d_i+d_i'$, and $q_o(s)$ and $u_o(s)$ are the reference linear and angular strains.
Given the tendon routing, $\Lambda_{\text{act}}(s,t)$ is fully determined by the tendon tensions $\tau_i(t)$. This model approximates tendon effects as a distributed wrench field, and it has been shown to provide a good approximation even for continuum robots with spacer disks \citep{rao2021model}.
Other actuation mechanisms, such as fluidic and pneumatic actuation, can be modeled within the same framework as~\eqref{eq:actuator model} under appropriate assumptions, including incompressible flow or quasi-static aerodynamics, and constant chamber cross-sectional area. In this case, $\tau_i$ denotes the fluidic or pneumatic forces generated by the actuating chambers~\citep{till2019real,renda2020geometric}.

Finally, we have the following boundary conditions at $s=0$ (the base) and $s=L$ (the tip):
\begin{align}
    g(0,t)=g_0(t), \eta(0,t)=\eta_0(t),  \Lambda(L,t)=\F_1(t),
\end{align}
where $g_0(t)$ is the base pose, $\eta_0(t)$ is the base velocity twist (zero for fixed base), and $\F_1(t)$ is the wrench applied at the tip (zero if unloaded).

From a control systems perspective, these equations represent the system model, where the tendon tensions $\tau_i(t)$ act as inputs. When supplied with initial conditions $\xi(s,0)$ and $\eta(s,0)$, the system becomes fully determined and one can simulate the system forward in time to compute all future states, a process called \textit{simulation} or \textit{model prediction}. Examples of model prediction for continuum robots include \citep{till2019real, mathew2022sorosim}. In practice, however, pure model prediction is often unreliable for several reasons. First, the initial conditions may not be accurately known, especially if the algorithm is initialized when the robot is already in motion or has undergone unknown perturbations. In fact, if we consider the robot's state immediately after a disturbance as the new initial condition for subsequent evolution, then the initial condition is practically never known. Second, modeling errors, numerical integration inaccuracies, and external disturbances inevitably accumulate and cause the predicted trajectory to diverge from the actual motion over time. This is why we need \textit{state estimation} \citep{choset2005PrinciplesRobotMotion}. The core idea is to fuse additional measurements, called \textit{outputs}, with model predictions to obtain more accurate estimates of the system's internal state, with the Kalman filter \citep{barfoot2017state} as a representative. Importantly, a state estimation algorithm does not require knowing the true initial condition.
 
Effective state estimation requires selecting outputs that are both easily measurable and sufficiently informative to recover unknown states. 
Measurements at the boundaries are generally more accessible than those within the body of the robot. For example, force/torque sensors can be readily mounted at the base, and motion capture markers and IMUs can be placed at the tip.
The corresponding estimation algorithms are thus called \textit{Cosserat Rod boundary observers}.
The boundary conditions at the base and tip highlight the dual nature of these observers. At the tip, the internal wrench is pre-specified, while the velocity twist varies and provides insights into the unknown dynamic states of the robot. This principle guided the development of a \textit{tip observer} in \citep{zheng2024estimating, zheng2024full}. Conversely, at the base, the velocity twist is pre-specified, and the internal wrench is a function of the robot's current state. This symmetry inspired the design of a dual \textit{base observer} in this work.

\subsection{Problem Statement}

Using the notation defined in the previous section, the problem can be stated as follows.

\textbf{Model and input.}
Assume the following are known: the coefficients $\M(s)$ and $\K(s)$, the boundary conditions $g_0(t)$, $\eta_0(t)$, and $\F_1(t)$, and the tendon inputs ${\tau_i(t)}$. Note that the initial conditions at $t=0$ are not needed.

\textbf{Measurement.}
Assume that we can measure the internal wrench at the base, denoted by $\bar{\Lambda}_0(t)$, and/or the pose and velocity at the tip, denoted by $\bar{g}_1(t)$ and $\bar{\eta}_1(t)$.

The objective of a state estimation problem is to estimate the \textit{infinite-dimensional} states of the soft continuum robot, including the poses $g(s,t)$, the strains $\xi(s,t)$, the velocities $\eta(s,t)$, and the internal wrenches $\Lambda(s,t)$.

Wrenches can be measured using embedded FT sensors, while rotation and angular velocities can be obtained from embedded IMUs. In contrast, position and linear velocity measurements typically rely on external motion capture systems. For outdoor or unstructured environments, embedded sensing is generally preferred.


\subsection{Special Cases}  \label{sec:Kirchhoff}

Before introducing the estimation algorithm, it is insightful to examine special cases where certain robot states are fully determined by others. These cases are particularly useful for state estimation, as they imply that some measurements inherently encode information about others. This observation can be leveraged to relax sensing requirements.

\subsubsection{1) Angular Motion Determines Linear Motion: Kirchhoff Rod.}

The full Cosserat rod kinematics have six internal degrees of freedom (DOFs): two bending strains, one torsional strain, two shear strains, and one elongation strain.
For continuum robots consisting of a slender elastic backbone, shear and elongation often have a negligible impact compared with bending and torsion, as the bending stiffness is proportional to $r^4$ while the shear stiffness is proportional to $r^2$, where $r$ is the radius of the rod.
By assuming constant linear strain, i.e., $q\equiv q_o$, the Cosserat rod reduces to a Kirchhoff rod with only three internal DOFs associated with the angular strain $u$.
In this case, the linear motion ($p$, $v$) is completely determined by the angular motion ($R$, $u$, $w$).
To see this, we collect all the equations for the Kirchhoff case.
\begin{align}
    R_t & = Rw\cross \label{eq:Rt} \\
    p_t & = Rv  \label{eq:pt}\\
    R_s & = Ru\cross \label{eq:Rs} \\
    p_s & = Rq \label{eq:ps} \\
    u_t & =w_s+u\cross w \label{eq:ut} \\
    q_t & =v_s+u\cross v-w\cross q \label{eq:qt} \\
    \M_a w_t & =m_s+u\cross m+q\cross n \nonumber \\
    & \quad - w\cross\M_a w - v\cross\M_l v+l \label{eq:wt} \\
    \M_l v_t & =n_s+u\cross n- w\cross\M_lv +f \label{eq:vt}
\end{align}
where $m=K_a(u-u_o)+m_{\text{act}}$ and $n$ now acts as a Lagrange multiplier to enforce $q\equiv q_o$.
We perform the following analysis at every fixed time instant $t$.
Using \eqref{eq:ps}, we see the position field $p$ is completely determined by the rotation field $R$ by
\begin{align}
    p(s) = p(0) + \int_0^s R(x)q_o(x) \,dx,
\end{align}
Using \eqref{eq:qt} and $q_t\equiv0$,
\begin{align}
    v_s+u\cross v & = w\cross q_o \\
    Rv_s+Ru\cross v & = Rw\cross q_o \\
    (Rv)_s & = Rw\cross q_o \\
    v(s) & = R^\top(s)\Big[R(0)v(0) \nonumber \\
    & \quad + \int_0^s R(x)w(x)\cross q_o(x) \,dx\Big], \label{eq:v vs w}
\end{align}
which suggests that $v$ is completely determined by $w$ and $R$.
Differentiating $\eqref{eq:v vs w}$ in $t$ reveals that $v_t$ is completely determined by $w_t$, $w$, and $R$. 

\subsubsection{2) Linear Motion Determines Angular Motion: Planar Kirchhoff Rod.}

Now it would be interesting to ask when the opposite is true. One such example is the planar Kirchhoff case. 
In this case, the angular motion $(R, w)$ is completely determined by the linear motion $(p, v)$. To see this, notice that in the planar case, $R$ can be parametrized by a scalar angle field $\theta(s)$ and \eqref{eq:ps} simplifies to
\begin{align}
    p_s = \begin{bmatrix}
        \cos(\theta) & -\sin(\theta) \\
        \sin(\theta) & \cos(\theta)
    \end{bmatrix}q_o,
\end{align}
which has a unique solution for $\theta(s)$ given $p_s(s)$.
Taking the time derivative of both sides shows that $(\theta,w)$ is completely determined by $(p,v)$.

The explicit redundancies revealed in this section may not be directly useful for simulation algorithms, as the dependencies often take complex forms. However, they are particularly valuable in the context of state estimation, as they allow for the relaxation of sensing requirements for states that are fully determined by others. This observation will be validated through simulation studies presented later.
In fact, we hypothesize that a similar redundancy may exist between the internal moment $m$ and internal force $n$. While their relationship is likely implicit and difficult to express analytically, we will provide supporting evidence through simulation and experimental results in subsequent sections.

\begin{remark}
The discussion in this section does not constitute a formal observability analysis in the control-theoretic sense. To the best of our knowledge, the most closely related work is~\cite{brace2022sensor}, which investigates observability for an Euler--Bernoulli beam, i.e., a linearized Kirchhoff rod. The results in~\cite{brace2022sensor} indicate that observability is maximized when curvature is measured at the fixed base. This observation is consistent with our findings, where internal moments (proportional to curvature) enable recovery of the full state in the Kirchhoff rod setting. A promising direction for future work is to analyze local observability via linearization and to extend the techniques developed in~\cite{brace2022sensor}.
\end{remark}

\section{Design of Cosserat Rod Boundary Observers}
\label{sec:algorithm}

In this section, we introduce the theoretical motivation behind our Cosserat rod boundary observer, present the algorithmic formulation, and describe its numerical implementation.

\subsection{Energy Dissipation of Cosserat Rods}
 
The design of our Cosserat rod boundary observers is inspired by the principle of energy dissipation. In a mechanical system with energy dissipation, the total energy tends to converge to zero over time. To illustrate this, consider a Cosserat rod without actuation and external distributed wrenches, i.e., $\Lambda_{\text{act}}=0$ and $\F=0$.
The governing equations become:
\begin{align} \label{eq:full governing equations}
\begin{split}
    \xi_t & =\eta_s+\ad_{\xi}\eta, \\
    \M\eta_t-\ad_\eta^\top\M\eta & =\Lambda_s-\ad_\xi^\top\Lambda,
\end{split}
\end{align}
where $\Lambda=\K(\xi-\xi_o)$.

The total energy of the system is given by:
\begin{align}
    E(t)=\frac{1}{2}\int_0^L(\xi-\xi_o)^\top\K(\xi-\xi_o)+\eta^\top\M\eta ds,
\end{align}
and its time derivative satisfies:
\begin{align*}
    \dot{E} & =\int_0^L\Lambda^\top\xi_t+\eta^\top\M\eta_t ds \\
    & =\int_0^L\Lambda^\top(\eta_s+\ad_\xi\eta)+\eta^\top(\Lambda_s-\ad_\xi^\top\Lambda+\ad_\eta^\top\M\eta) ds \\
    & =\Lambda^\top\eta\mid_0^L+\int_0^L\Big[-\Lambda_s^\top\eta+\Lambda^\top\ad_\xi\eta+\eta^\top\Lambda_s \\
    & \quad -(\ad_\xi\eta)^\top\Lambda+(\ad_\eta\eta)^\top\M\eta\Big] ds \\
    & =\Lambda^\top(L,t)\eta(L,t)-\Lambda^\top(0,t)\eta(0,t),
\end{align*}
where the third equality follows from applying integration by parts to the term $\Lambda^\top\eta_s$.

The expression for $\dot{E}$ shows how the total energy is influenced by energy exchange through the boundaries. At the tip ($s=L$), if we define the point wrench as $\Lambda(L,t)=-\Gamma_1\eta(L,t)$, where $\Gamma_1\in\mathbb{R}^{6\times6}$ and $\Gamma_1\succ0$, we obtain a negative term $-\eta^\top(L,t)\Gamma_1\eta(L,t)$.\footnote{We use $\succ0$ and $\succeq0$ to represent positive-definite and positive semidefinite matrices.} The physical implication is that a damping wrench is applied at the tip, which dissipates the system's total energy. This principle motivates the design of the tip boundary observer in \citep{zheng2024estimating, zheng2024full}.

Similarly, at the base ($s=0$), if we set the base velocity as $\eta(0,t)=\Gamma_0\Lambda(0,t)$, where $\Gamma_0\succ0$, we obtain another negative term, $-\Lambda^\top(0,t)\Gamma_0\Lambda(0,t)$. The physical implication is that the base velocity ``swings'' the rod in a way that dissipates the total energy, which motivates the design of the base boundary observer in this work.


\subsection{Dissipation-Inspired Boundary Observers}
\label{sec:observer design}

Inspired by the energy dissipation properties of the ``swinging'' base, we developed a novel base observer algorithm to estimate all robot states.
The core equations are:
\begin{align}
    \xi_t & =\eta_s+\ad_{\xi}\eta, \label{eq:boundary observer 1} \\ 
    \M\eta_t-\ad_\eta^\top\M\eta & =\Lambda_s-\ad_\xi^\top\Lambda+\F, \label{eq:boundary observer 2}
\end{align}
with supplementary equations \eqref{eq:forward kinematics} and  \eqref{eq:external wrench}-\eqref{eq:actuator model} to construct the necessary intermediate variables $g(s,t)$, $\F(s,t)$, $\Lambda(s,t)$, and $\Lambda_{\text{act}}(s,t)$.
These equations essentially mirror those of the Cosserat rod theory. The novelty lies in the modified boundary conditions.
For the base observer, the modified boundary conditions are given by:
\begin{align}
    & g(0,t)=g_0(t), \label{eq:base observer BC 1} \\ 
    & \eta(0,t)=\eta_0(t) + \Gamma_0\big(\Lambda(0,t)-\bar{\Lambda}_0(t)\big), \label{eq:base observer BC 2} \\ 
    & \Lambda(L,t)=\F_1(t), \label{eq:base observer BC 3}
\end{align}
where $g_0(t)$ and $\eta_0(t)$ are the known physical base pose and velocity, $\F_1(t)$ is the known physical tip wrench, $\Lambda(0,t)$ is the current estimated internal wrench at the base computed by the observer algorithm, and $\bar{\Lambda}_0(t)$ is the measured internal wrench from a 6-axis FT sensor.
This base observer contrasts with the tip observer introduced in \citep{zheng2024estimating,zheng2024full}, differing only in the boundary conditions.

Note that due to the injected virtual velocity at the base, $\Gamma_0\big(\Lambda(0,t)-\bar{\Lambda}_0(t)\big)$, the observer's output represents a ``flying'' robot whose configuration converges to the actual robot but may exhibit a different base frame. However, the boundary condition \eqref{eq:base observer BC 1} ensures that the estimated configuration, $g(s,t)$, is always reconstructed from the physical base frame, $g_0(t)$.
Additionally, the estimated velocity, $\eta(s,t)$, is also influenced by the virtual velocity at the base. To eliminate this artifact, we can recompute $\eta(s,t)$ directly from the reconstructed configuration $g(s,t)$.

For completeness, we also present the boundary conditions of the tip observer:
\begin{align}
    & g(0,t)=g_0(t), \label{eq:tip observer BC 1} \\
    & \eta(0,t)=\eta_0(t), \label{eq:tip observer BC 2} \\
    & \Lambda(L,t)=\F_1(t)-\Gamma_1\big(\eta(L,t)-\bar{\eta}_1(t)\big), \label{eq:tip observer BC 3}
\end{align}
where $\eta(L,t)$ is the current estimate of the tip velocity and $\bar{\eta}_1(t)$ is the measured tip velocity. The tip observer does not exhibit the artifact of ``flying'' configurations because the injected term is a wrench.

The added terms $\Gamma_0\big(\Lambda(0,t)-\bar{\Lambda}_0(t)\big)$ and $-\Gamma_1\big(\eta(L,t)-\bar{\eta}_1(t)\big)$ are referred to as \textit{boundary correction terms}. The first acts as a virtual swinging base, and the second as a virtual damping wrench at the tip. It is important to note that these correction terms dissipate the energy associated with the estimation errors rather than the actual robot system. By subtracting the equations for the estimated states from those of the actual robot, one can see that the correction terms act as dissipative terms in the error dynamics.


The most general boundary conditions of a boundary observer are given by:
\begin{align}
    g(0,t) & =g_0(t), \label{eq:boundary observer BC 1} \\
    \eta(0,t) & =\eta_0(t) + \Gamma_0\big(\Lambda(0,t)-\bar{\Lambda}_0(t)\big), \label{eq:boundary observer BC 2} \\
    \begin{split} \label{eq:boundary observer BC 3}
        \Lambda(L,t) & =\F_1(t)-\Gamma_{\text{P}}\big[\log \big(\bar{g}_1^{-1}(t)g(L,t)\big)\big]\ve \\
        & \quad -\Gamma_{\text{D}}\big(\eta(L,t)-\bar{\eta}_1(t)\big),
    \end{split}
\end{align}
where $g(L,t)$ is the current estimate of the tip pose, $\bar{g}_1(t)$ is the measured tip pose, ``$\vee$'' (defined in the Appendix) converts a $se(3)$ twist into a 6-vector, and $\Gamma_{\text{P}},\Gamma_{\text{D}}\succ0$.
The subscripts ``P'' and ``D'' indicate their similarity to a proportional-derivative controller.
In the case of tip observers, depending on whether the P term is present, we will refer to the corresponding observer as \textit{tip D observer} and \textit{tip PD observer}.


As explained before, in certain special cases, some states are fully determined by others. In such scenarios, the corresponding correction terms in \eqref{eq:boundary observer BC 1}--\eqref{eq:boundary observer BC 3} can be removed. This simplification is valid when the model and external inputs are fully known. In the presence of significant modeling uncertainties or unknown inputs, correction terms based on base wrench or tip velocity can struggle to eliminate steady-state errors. In this case, the inclusion of the proportional term helps mitigate this issue by generating additional corrective inputs. These properties will be demonstrated through simulation-based and experimental validation in the subsequent sections.

\begin{remark}
Boundary state estimation for one-dimensional PDEs has been extensively studied~\citep{krstic2008boundary,vazquez2011backstepping,castillo2013boundary}. Existing boundary observer designs typically assume linear dynamics or, for certain classes of nonlinear PDEs, rely on nontrivial integral transformations to recast the system into a linear form, or impose global Lipschitz conditions on the nonlinear terms. 
In contrast, the proposed method directly addresses a nonlinear PDE with Lie group structure. By leveraging an energy-based design, we obtain a conceptually simple observer with guaranteed convergence.
\end{remark}

\section{Optimal Boundary Observers}

The estimation errors converge as long as the observer gains are positive-definite. In this section, we aim to investigate the choice of \textit{optimal gains} that yield the fastest possible convergence.
However, the governing equations~\eqref{eq:full governing equations} are highly nonlinear due to the underlying Lie group structure, making it extremely challenging to derive explicit expressions for the optimal observer gains. Notably, all the Lie bracket terms (the nonlinear terms involving the adjoint operator) arise from the fact that the robot configuration $g(s,t)$ evolves on the Lie group $SE(3)$.
To address this challenge, we consider a \textit{linear approximation} of the governing equations. Note that this is not a standard linearization about a trajectory or equilibrium point. Rather, we assume from the beginning that $g(s,t)$ lies in a Euclidean space $\mathbb{R}^n$. Under this assumption, following the same procedure to define the remaining states yields
\begin{align}   \label{eq:g Euclidean}
\begin{split}
    g_t=\eta,  \\
    g_s=\xi. 
\end{split}
\end{align}
The equality of mixed partial derivatives $\partial_{st}g=\partial_{ts}g$ yields the compatibility equation:
\begin{align}
    \xi_t=\eta_s,
\end{align}
where we notice that the Lie bracket has disappeared.
Similarly, the dynamics equation would then take a much cleaner form:
\begin{align}
    \M\eta_t=\Lambda_s,
\end{align}
where $\Lambda=K(\xi-\xi_o)$.
Again, the two Lie brackets disappear because Euclidean structure is assumed from the outset.
The approximated governing system becomes
\begin{align}   \label{eq:linear governing equations}
\begin{aligned}
    \xi_t       & = \eta_s, \\
    M\eta_t     & = \Lambda_s,
\end{aligned}
\qquad
\text{or}
\qquad
\begin{aligned}
    \Lambda_t   & = \K\eta_s, \\
    \eta_t      & = \M^{-1}\Lambda_s,
\end{aligned}
\end{align}
with dissipative boundary conditions
\begin{align}   \label{eq:eq:linear governing equations BC}
\begin{split}
    \eta(0,t)       & = \Gamma_0\Lambda(0,t), \\
    \Lambda(L,t)    & = -\Gamma_1\eta(L,t),
\end{split}
\end{align}
where $\Gamma_0$ and $\Gamma_1$ are positive definite.
Notably, the approximated system \eqref{eq:linear governing equations} is a \textit{linear hyperbolic system}, whereas the original system~\eqref{eq:full governing equations} is a \textit{semilinear hyperbolic system}. In the PDE literature, the local qualitative behavior of a PDE is typically governed by its highest-order derivatives \citep{evans1988partial}. Therefore, it is reasonable to analyze the local asymptotic behavior of~\eqref{eq:full governing equations} using its linear counterpart~\eqref{eq:linear governing equations}, especially because any Lie group is locally diffeomorphic to a Euclidean space of the same dimension~\citep{lee2003smooth}. The validity of this approximation is supported by the simulation studies presented later.

\begin{remark}
The linearization is introduced solely to facilitate the analysis of the existence of an optimal observer gain. The observer design and the energy-based convergence analysis established in the previous sections do not rely on this linearization.
\end{remark}

\subsection{Riemann Coordinates and Canonical Forms}

To analyze the convergence behavior, a standard procedure is to convert \eqref{eq:linear governing equations} into a canonical form through a coordinate transformation~\citep{bastin2016stability}. Since $M$ and $K$ are symmetric and positive definite, define
\begin{align}
    S := K^{\frac{1}{2}} M^{-1} K^{\frac{1}{2}}.
\end{align}
Then $S$ is symmetric positive definite and admits a diagonalization
\begin{align}   \label{eq:Sigma}
    S = U^\top \Sigma^2 U, 
\end{align}
where $U^\top U = I$ and $\Sigma$ is a diagonal matrix consisting of the square roots of the eigenvalues of $S$.
We can now define the \textit{Riemann coordinates} through the coordinate transformation
\begin{align}   \label{eq:phi-def}
\begin{split}
    \phi_+  & =U \K^{\frac{1}{2}}\eta - \Sigma U \K^{-\frac{1}{2}}\Lambda,  \\
    \phi_-  & =U \K^{\frac{1}{2}}\eta + \Sigma U \K^{-\frac{1}{2}}\Lambda.
\end{split}
\end{align}
The inverse transform is given by:
\begin{align}   \label{eq:inverse transform}
\begin{split}
    \Lambda &= \frac{1}{2}\K^{\frac{1}{2}} U^\top \Sigma^{-1} (\phi_- - \phi_+),    \\
    \eta    &= \frac{1}{2}\K^{-\frac{1}{2}} U^\top (\phi_- + \phi_+).
\end{split}
\end{align}
By substituting the above transform into \eqref{eq:linear governing equations}, the Riemann coordinates satisfy the following characteristic form:
\begin{align}   \label{eq:Riemann}
\begin{split}
    (\phi_+)_t + \Sigma (\phi_+)_s=0,   \\
    (\phi_-)_t - \Sigma (\phi_-)_s=0.
\end{split}
\end{align}
The equivalent representation above is known as the canonical \textit{Riemann form}. A key advantage of this formulation is that the transformed state variables $\phi_+$ and $\phi_-$ are decoupled in the governing equation \eqref{eq:Riemann}, in contrast to the original form \eqref{eq:linear governing equations}, where they are intrinsically coupled. In this Riemann form, any coupling between $\phi_+$ and $\phi_-$ is only through the boundary conditions.
To obtain the boundary condition for the Riemann form \eqref{eq:Riemann}, we can substitute \eqref{eq:inverse transform} into \eqref{eq:eq:linear governing equations BC} and obtain
\begin{align}   \label{eq:Riemann BC}
\begin{split}
    \phi_+(0,t) & = \rho_0\phi_-(0,t),   \\
    \phi_-(L,t) & = \rho_1\phi_+(L,t),
\end{split}
\end{align}
where
\begin{align}   \label{eq:rho0-rho1}
\begin{split}
    \rho_0  & = \big(\Sigma^{-1} + G_0\big)^{-1}\big(\Sigma^{-1} - G_0\big),   \\
    & = (I + \Sigma G_0)^{-1}(I - \Sigma G_0),  \\
    \rho_1  & = \big(\Sigma^{-1} + G_1\big)^{-1}\big(\Sigma^{-1} - G_1\big),    \\
    & = (I + \Sigma G_1)^{-1}(I - \Sigma G_1)
\end{split}
\end{align}
and
\begin{align}   \label{eq:G0-G1}
\begin{split}
    G_0 & = U\K^{-\frac{1}{2}}\Gamma_0^{-1}\K^{-\frac{1}{2}}U^\top,  \\
    G_1 & = U\K^{-\frac{1}{2}}\Gamma_1\K^{-\frac{1}{2}}U^\top.
\end{split}
\end{align}
The boundary conditions \eqref{eq:Riemann BC} for the Riemann form have an intuitive physical interpretation. By substituting \eqref{eq:g Euclidean} into the second equation of \eqref{eq:linear governing equations}, we obtain
\begin{align}
    g_{tt} = M^{-1}\big(K(g_s-\xi_o)\big)_s,
\end{align}
which is known as a \textit{wave equation}, commonly used to model, e.g., string vibrations.
Then, in \eqref{eq:Riemann BC}, $\rho_0$ represents how much of the outgoing left-moving wave $\phi_-$ is reflected and converted into the incoming right-moving wave $\phi_+$, and similarly for $\rho_1$. 
When $M$ and $K$ depend on $s$, there will be additional zeroth-order terms in \eqref{eq:Riemann}. However, since our analysis is based on an approximation using only the highest order derivatives, we would again drop the additional terms.

\subsection{Estimate of Convergence Rate}

Next, we need to extend the Lyapunov analysis in \citep{bastin2016stability} Chapter 2 from the scalar case to the matrix case to estimate the decay rate. 

\medskip
\noindent\textbf{Lyapunov functional.} Fix $\mu>0$ and define the weights
\begin{align}
    W_-(s):=e^{+\mu s\Sigma^{-1}},  \quad
    W_+(s):=e^{-\mu s\Sigma^{-1}}.
\end{align}
Then $W_\pm(s)\succ0$ and $[W_\pm(s),\Sigma]=0$, where $[A,B]:=AB-BA$ denotes the commutator. Their derivatives satisfy
\begin{align}   \label{eq:Ws}
    W_-'(s)=+\mu\Sigma^{-1}W_-(s),  \,
    W_+'(s)=-\mu\Sigma^{-1}W_+(s). 
\end{align}
Let $P_\pm\succ0$ be symmetric matrices whose values will be chosen later, and require $[P_\pm,\Sigma]=0$.
We consider the Lyapunov functional
\begin{align}    \label{eq:Vdef}
    V(t) =\int_0^L\Big[ 
    \phi_-^\top P_-\Sigma^{-1}W_-\phi_-   
    + \phi_+^\top P_+\Sigma^{-1}W_+\phi_+
    \Big]ds.
\end{align}
Then, $V$ is equivalent to $\|\phi_-\|_{L^2}^2+\|\phi_+\|_{L^2}^2$.

\medskip
\noindent\textbf{Time derivative.}
Differentiating \eqref{eq:Vdef} in time (note $W_\pm$ do not depend on $t$) and using \eqref{eq:Riemann},
\begin{align*}
    \dot V(t)
    & = 2\int_0^L\Big[
    \phi_-^\top P_-\Sigma^{-1}W_-(\phi_-)_t  \\
    & \qquad \qquad \qquad +\, \phi_+^\top P_+\Sigma^{-1}W_+(\phi_+)_t
    \Big]ds\\
    & = 2\int_0^L\Big[
    \phi_-^\top P_-W_-(\phi_-)_s  - \phi_+^\top P_+W_+(\phi_+)_s
    \Big]ds.
\end{align*}
Applying the identity
\begin{align*}
    2\phi^\top A\phi_s=\big(\phi^\top A \phi\big)_s - \phi^\top A_s \phi
\end{align*}
to each integral (with $A=P_\pm W_\pm$) yields
\begin{align*}
    \dot V(t)
    & = \Big[\phi_-^\top P_-W_-\phi_-\Big]_{0}^{L}
    -\int_0^L\phi_-^\top P_-W_-'\phi_-ds \\
    & \quad -\Big[\phi_+^\top P_+W_+\phi_+\Big]_{0}^{L}
    +\int_0^L\phi_+^\top P_+W_+'\phi_+ds.
\end{align*}
Using \eqref{eq:Ws} we obtain 
\begin{align}
\begin{split}
    \dot V(t)
    & = \Big[\phi_-^\top P_-W_-\phi_-\Big]_{0}^{L}
    -\Big[\phi_+^\top P_+W_+\phi_+\Big]_{0}^{L} -\mu V(t).
\end{split}
\end{align}

\medskip
\noindent\textbf{Boundary terms.}
At $s=0$, $W_\pm(0)=I$ and by \eqref{eq:Riemann BC},
\begin{align*}
\begin{split}
    -\phi_-^\top P_-W_-\phi_-
    +\phi_+^\top P_+W_+\phi_+ = \phi_-^\top\big(\rho_0^\top P_+\rho_0 - P_-\big)\phi_-,
\end{split}
\end{align*}
which is nonpositive if
\begin{align} \label{eq:LMI0}
    \rho_0^\top P_+\rho_0\preceq P_-.
\end{align}
At $s=L$, by \eqref{eq:Riemann BC},
\begin{align*}
\begin{split}
    & \quad \phi_-^\top P_-W_-\phi_- - \phi_+^\top P_+W_+\phi_+ \\
    & = \phi_+^\top\big(\rho_1^\top P_-W_-\rho_1 - P_+W_+\big)\phi_+,
\end{split}
\end{align*}
which is nonpositive if
\begin{align} \label{eq:LMI1}
    \rho_1^\top P_-e^{\mu L\Sigma^{-1}}\rho_1
    \preceq P_+e^{-\mu L\Sigma^{-1}}.
\end{align}

\medskip
\noindent\textbf{Exponential convergence.}
Therefore, a sufficient condition for $\dot V\le-\mu V$ is
\begin{align} \label{eq:LMI}
\left\{
\begin{aligned}
    &P_+,P_-\succ0,\qquad [P_\pm,\Sigma]=0,\\
    &\rho_0^\top P_+\rho_0\preceq P_-,\\
    &\rho_1^\top P_-e^{\mu L\Sigma^{-1}}\rho_1
      \preceq P_+e^{-\mu L\Sigma^{-1}}.
\end{aligned}
\right.
\end{align}
Accordingly, define the fastest energy decay rate certified by this family of Lyapunov functionals as
\begin{align} \label{eq:mu-max}
    \mu_{\max}:=\sup\{\mu>0:\ \eqref{eq:LMI}\text{ is feasible}\}.
\end{align}
For a fixed $\mu$, these conditions can be checked numerically. 

For the reflection matrices \eqref{eq:rho0-rho1}, with $G_i \succ 0$, define
$H_i=\Sigma^{1/2}G_i\Sigma^{1/2}\succ0$. Then
\begin{align*}
    \rho_i
    =\Sigma^{1/2}(I+H_i)^{-1}(I-H_i)\Sigma^{-1/2},
\end{align*}
and hence
\begin{align*}
    \Sigma^{-1}-\rho_i^\top\Sigma^{-1}\rho_i
    =4\Sigma^{-1/2}H_i(I+H_i)^{-2}\Sigma^{-1/2}\succ0.
\end{align*}
Therefore,
\begin{align*}
    \rho_i^\top\Sigma^{-1}\rho_i\prec\Sigma^{-1}.
\end{align*}
Thus, $P_+=P_-=\Sigma^{-1}$ satisfies the boundary inequalities strictly
at $\mu=0$, and by continuity \eqref{eq:LMI} remains feasible for
sufficiently small $\mu>0$.

\begin{figure}
    \centering
    \includegraphics[width=0.9\linewidth]{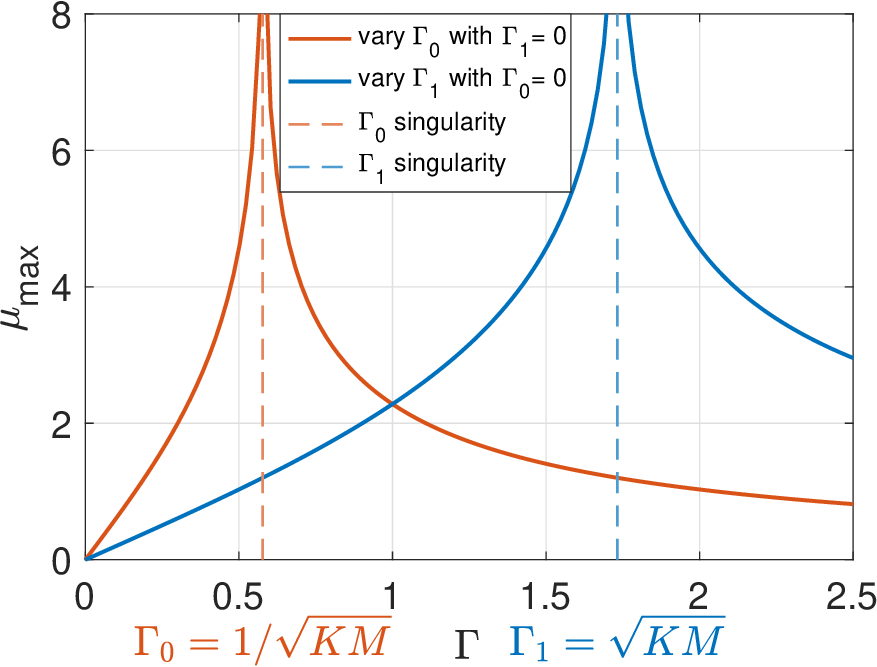}
    \caption{In the scalar case, the fastest convergence rate $\mu_{\max}$ versus $\Gamma_1$ (with $\Gamma_0=0$) and versus $\Gamma_0$ (with $\Gamma_1=0$). As $\Gamma_0$ or $\Gamma_1$ increases, $\mu_{\max}$ increases to infinity at a singularity and then decreases. System parameters: $L=1$, $M=1$, $K=3$.}
    \label{fig:mu-max}
\end{figure}

In the scalar case, \eqref{eq:LMI} is feasible if and only if
\begin{align}
    \rho_0^2\rho_1^2e^{2\mu L/\Sigma}\le1.
\end{align}
Consequently, the largest state-norm decay rate is
\begin{align}
    \alpha_{\max}:=\frac{\mu_{\max}}{2}=\frac{\Sigma}{2L}\ln\!\left(\frac{1}{|\rho_0\rho_1|}\right).
\end{align}
This expression is consistent with the scalar case in \citep{bastin2016stability} Chapter 2. An illustration is given in Fig.~\ref{fig:mu-max}, where holding one gain fixed and increasing the other causes $\mu_{\max}$ to increase to infinity at a singularity and then decrease.
The fastest convergence is achieved at the singularity, corresponding to the optimal gain.

\subsection{Optimal Gains and Finite-Time Convergence}  \label{sec:optimal gain}

To determine the optimal gains, it is interesting to notice that \eqref{eq:LMI} is feasible for any $\mu>0$ if $\rho_0=0$ or $\rho_1=0$.
Indeed, if $\rho_0=0$, the first boundary inequality is automatically satisfied, and for any fixed $\mu>0$ the second can be satisfied by choosing $P_-$ sufficiently small relative to $P_+$. Similarly, if $\rho_1=0$, the second boundary inequality is automatically satisfied, while the first is feasible, e.g., with $P_+=P_-=\Sigma^{-1}$.
The corresponding boundary is called ``perfectly absorbing''. In this case, it yields finite-time extinction, which is faster than exponential convergence.
If at least one boundary is perfectly absorbing, every wave packet disappears after at most one reflection. Consequently, the solution vanishes in
\begin{align}
    t\le \frac{2L}{\sigma_{\min}(\Sigma)},
\end{align}
corresponding to the maximum time for a wave to travel to a non-absorbing boundary, reflect once, and then reach the absorbing boundary. If both boundaries are absorbing, no reflection occurs and extinction occurs faster, within
\begin{align}
    t\le \frac{L}{\sigma_{\min}(\Sigma)}.
\end{align}
Here, the singular values of $\Sigma$ correspond to the characteristic wave speeds.
The improved convergence time of dual absorbing boundaries suggests the use of dual boundary observer techniques for faster state estimation.

Using \eqref{eq:rho0-rho1} and \eqref{eq:G0-G1} and setting $\rho_0=\rho_1=0$, the theoretical optimal gains are given by
\begin{align}   \label{eq:ref gain}
\begin{split}
    \Gamma_0'    & = K^{-\frac{1}{2}}\big(K^{\frac{1}{2}} M^{-1} K^{\frac{1}{2}}\big)^{\frac{1}{2}} K^{-\frac{1}{2}}, \\
    \Gamma_1'    & = K^{\frac{1}{2}}\big(K^{-\frac{1}{2}} M K^{-\frac{1}{2}}\big)^{\frac{1}{2}} K^{\frac{1}{2}} = (\Gamma_0')^{-1}.
\end{split}
\end{align}
When the robot is a cylinder or a tube, both $M$ and $K$ are diagonal, and the above simplifies to
\begin{align}   \label{eq:ref gain diagonal}
\begin{split}
    \Gamma_0' = (MK)^{-\frac{1}{2}}, \quad
    \Gamma_1' = (MK)^{\frac{1}{2}}.
\end{split}
\end{align}

The Lyapunov-based analysis provides reference values for selecting the observer gains. The actual convergence time is typically rescaled by a factor dependent on the numerical scheme employed to solve the PDEs. 
Although the analysis in this section is based on a linear approximation, subsequent simulation studies demonstrate that the actual optimal gains quantitatively align with the reference values for a ``balanced'' Cosserat rod when the elements of $\M$ and $\K$ are of the same order. 
In the case of Kirchhoff rods, the maximum and minimum singular values of $\K$ or $\M$ can differ by the magnitude of $1/r^2$ (typically between $10^4$ and $10^6$). Simulation and experimental studies indicate that the actual optimal gains are also rescaled from the reference values, likely due to the non-negligible influence of nonlinear terms. Nevertheless, the overall qualitative trend remains consistent: increasing the observer gains initially reduces the convergence time, reaching an optimal value, beyond which further increase leads to slower convergence. This convex behavior enables an efficient and straightforward search for the optimal gains.

\section{Numerical Implementation}
\label{sec:numerical implementation}

\begin{algorithm}[!htbp]
\SetAlgoLined
\LinesNumbered
\caption{Shooting-Based Implementation of Cosserat Rod Boundary Observers}
\label{alg:boundary observer}
\KwIn{\\
    Initial conditions: $\xi(s,0)$, $\eta(s,0)$ \\
    Boundary conditions: $g_0(t)$, $\eta_0(t)$, $\F_1(t)$ \\
    Robot coefficients: $\M(s)$, $\K(s)$ \\
    Tendon tensions: $\tau_i(t)$ \\
    Boundary measurements: $\bar{\Lambda}_0(t)$, $\bar{g}_1(t)$, $\bar{\eta}_1(t)$ \\
}
\KwOut{\\
    State estimates: $\xi(s,t)$, $\eta(s,t)$, $g(s,t)$, $\Lambda(s,t)$ \\
}

\For{$t = 0:T$}{
    Residual error: $\gamma=1$ \\
    \While{$\gamma > \epsilon$}{
        Guess $\Lambda(0,t)$ \\
        \If{$\bar{\Lambda}_0(t)$ available}{
            Implement base correction using \eqref{eq:boundary observer BC 2}: \\
            $\eta_0(t) = \eta_0(t) + \Gamma_0\big(\Lambda(0,t)-\bar{\Lambda}_0(t)\big)$
        }
        \For{$s = 0:L$}{
            Calculate implicit approximation of time derivatives: $\xi_t$, $\eta_t$ \\
            Implement spatial integration for \eqref{eq:boundary observer 1}-\eqref{eq:boundary observer 2} \\
        }
        \If{$\bar{g}_1(t)$, $\bar{\eta}_1(t)$ available}{
           Implement tip correction using \eqref{eq:boundary observer BC 3}: \\
           $\F_1(t) = \F_1(t)-\Gamma_{\text{P}}\big[\log \big(\bar{g}_1^{-1}(t)g(L,t)\big)\big]\ve$ \\ 
           \qquad\qquad $- \Gamma_{\text{D}}\big(\eta(L,t)-\bar{\eta}_1(t)\big)$
        }
        Update the residual error: $\gamma=\|\Lambda(L,t)-\F_1(t)\|$
    }
    Recompute $\eta(s,t)$ from $g(s,t)$
}
\end{algorithm}

\begin{algorithm}[!htbp]
\SetAlgoLined
\LinesNumbered
\caption{Geometric Bracketing Search for Gain Scaling}
\label{alg:gain tuning}
\KwIn{Theoretical gain $\Gamma'$, initial scale $\gamma_c>0$, ratio $r>1$, samples $N$, tolerance $\varepsilon$}
\KwOut{Optimal scaling $\gamma^\star$}

\While{$r-1>\varepsilon$}{
    $\gamma^{(i)} = \gamma_c \, r^{\,i-\lceil N/2\rceil},\; i=1,\dots,N$\;
    
    \For{$i=1:N$}{
        Evaluate convergence time $J(\gamma^{(i)})$ using $\Gamma=\gamma^{(i)}\Gamma'$\;
    }
    
    $k=\arg\min_i J(\gamma^{(i)})$\;
    
    \eIf{$k=1$}{
        $\gamma_c \leftarrow \gamma^{(1)}$ \tcp*{shift left}
    }{
        \eIf{$k=N$}{
            $\gamma_c \leftarrow \gamma^{(N)}$ \tcp*{shift right}
        }{
            $\gamma_c \leftarrow \gamma^{(k)},\quad r \leftarrow \sqrt{r}$ \tcp*{refine}
        }
    }
}
\Return{$\gamma^\star=\gamma_c$}
\end{algorithm}

In our previous work \citep{zheng2024full, zheng2024estimating}, the tip observer was implemented using a strain parametrization \citep{boyer2020dynamics} that did not operate in real time.
In this section, we introduce a new numerical implementation based on shooting methods, one of the most efficient simulation methods for Cosserat rod dynamics, to achieve real-time state estimation. Readers unfamiliar with shooting methods are referred to \citep{till2019real, janabi2021cosserat, boyer2022statics}. In brief, shooting methods are widely used for solving two-point boundary value problems and involve two integration loops. At each fixed time step $t$, the forward dynamics of the Cosserat rod equations \eqref{eq:boundary observer 1}-\eqref{eq:boundary observer 2}, where all time derivatives are treated as given coefficients, form a system of ordinary differential equations (ODEs) in $s$. This constitutes a two-point boundary value problem, as the boundary conditions for $\eta$ in \eqref{eq:boundary observer 1} are specified at $s=0$, while those for $\Lambda$ in \eqref{eq:boundary observer 2} are given at $s=L$. 

Shooting methods involve an initial guess for an additional boundary condition, such as $\Lambda$ at $s=0$, followed by integrating the ODE system from $s=0$ to $s=L$ to compute the residual error for $\Lambda$ at $s=L$. This process iterates until convergence. The resulting shooting process generates time derivatives $\xi_t$ and $\eta_t$, which are then utilized for temporal integration. In practice, implicit temporal integration is often preferred for improved accuracy and numerical stability.

In Algorithm~\ref{alg:boundary observer}, we provide pseudocode for implementing boundary observers using shooting methods. Note that the initial conditions in this algorithm can be arbitrary initial guesses. Apart from the two ``if'' statements, the remainder of the algorithm is equivalent to the shooting method described in~\citep{till2019real}. This demonstrates that implementing the proposed boundary observer requires only a minor modification of the boundary conditions within an existing simulation framework.
Moreover, the conditional statements ensure that if a measurement is unavailable during a given iteration, the algorithm simply proceeds using the nominal Cosserat rod dynamics. As shown in~\citep{till2019real}, the computational complexity of the shooting method scales with the number of spatial and temporal discretization nodes. Real-time performance with 300 spatial nodes at 200~Hz has been demonstrated using an optimized C++ implementation on pre-2019 hardware (see Fig.~3 in~\citep{till2019real}). The impact of discretization levels on computational cost will be further evaluated in subsequent sections.

We also present a bracketing search algorithm (Algorithm~\ref{alg:gain tuning}) to determine the optimal gains by tuning scalar scaling factors applied to the theoretical gains. The convergence time is evaluated over a geometric grid of scaling factors to identify a candidate interval. If the objective is monotonic, the interval is shifted in the descent direction; otherwise, it is refined around the minimum using a coarse-to-fine strategy.

Let $\Gamma_{0}', \Gamma_{1}'$ denote the theoretical gains computed from \eqref{eq:ref gain}, with
\begin{align*}
    \Gamma_{0}' &= \diag(\Gamma_{0m}', \Gamma_{0n}'), \qquad
    \Gamma_{1}' = \diag(\Gamma_{1w}', \Gamma_{1v}'),
\end{align*}
where $\Gamma_{0m}', \Gamma_{0n}', \Gamma_{1w}', \Gamma_{1v}' \in \mathbb{R}^{3\times3}$ correspond to the gains for internal moment, internal force, angular velocity, and linear velocity, respectively.

\textit{Kirchhoff case:} The gains are scaled block-wise as
\begin{align*}
    \Gamma_{0m} &= \gamma_m \Gamma_{0m}',  &\quad \Gamma_{1w} &= \gamma_w \Gamma_{1w}', \\
    \Gamma_{0n} &= \gamma_n \Gamma_{0n}',  &\quad \Gamma_{1v} &= \gamma_v \Gamma_{1v}'.
\end{align*}
The scaling factors $\gamma_m,\gamma_n,\gamma_w,\gamma_v>0$ are tuned sequentially using Algorithm~\ref{alg:gain tuning}, with all other factors set to zero, to minimize convergence time. Planar simulations are used for this step, as each individual gain suffices to ensure convergence in the planar Kirchhoff setting. After tuning, the gains are assembled as
\begin{align*}
    \Gamma_{0} &= \diag(\Gamma_{0m}, \Gamma_{0n})/2, \qquad
    \Gamma_{1} = \diag(\Gamma_{1w}, \Gamma_{1v})/2.
\end{align*}
where the factor $1/2$ accounts for the fact that the linear and angular components carry equivalent information. 

\textit{Cosserat case:} The gains are scaled as
\begin{align*}
    \Gamma_{0} &= \gamma_0 \Gamma_{0}', \qquad
    \Gamma_{1} = \gamma_1 \Gamma_{1}',
\end{align*}
where $\gamma_0,\gamma_1>0$ are tuned sequentially using Algorithm~\ref{alg:gain tuning}.

Since the tuning procedure is identical in all cases, Algorithm~\ref{alg:gain tuning} is presented for a generic theoretical gain $\Gamma'$ and scaling factor $\gamma$.

\section{Simulation Study}
\label{sec:simulation}

\begin{figure*}[htbp]
    \centering
    \begin{subfigure}[b]{0.3\textwidth}
        \centering
        \includegraphics[height=4.5cm]{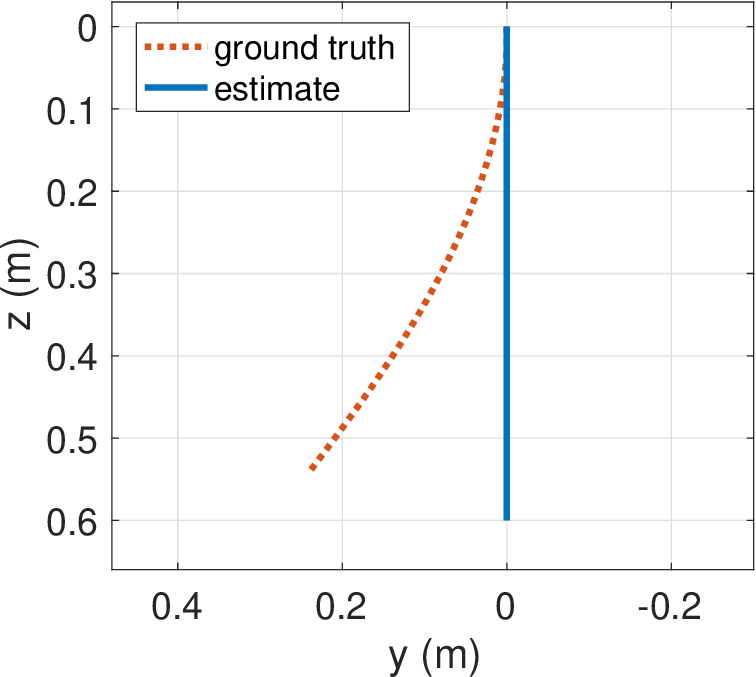}
    \end{subfigure}
    \hfill
    \begin{subfigure}[b]{0.34\textwidth}
        \centering
        \includegraphics[height=4.5cm]{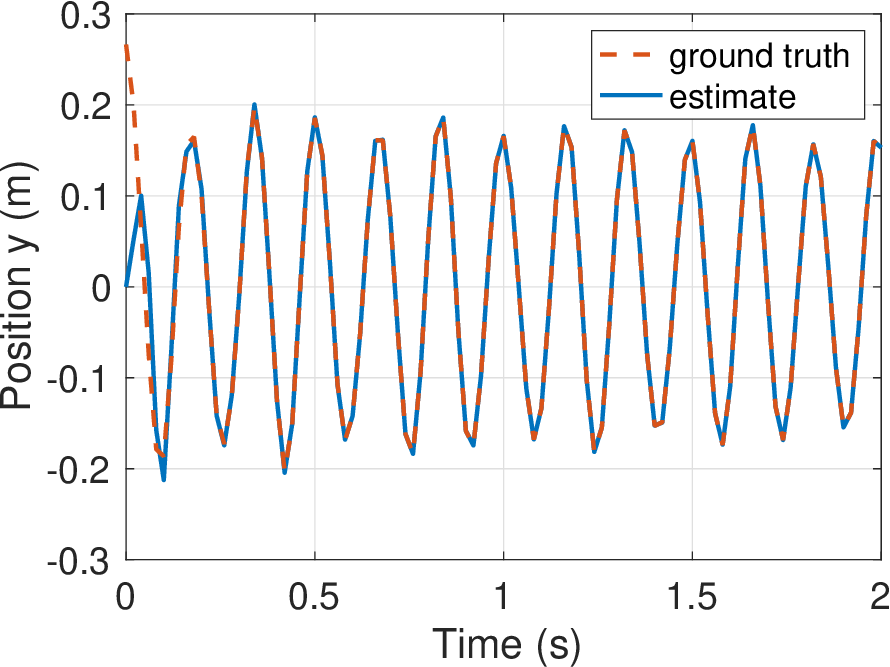}
    \end{subfigure}
    \hfill
    \begin{subfigure}[b]{0.34\textwidth}
        \centering
        \includegraphics[height=4.5cm]{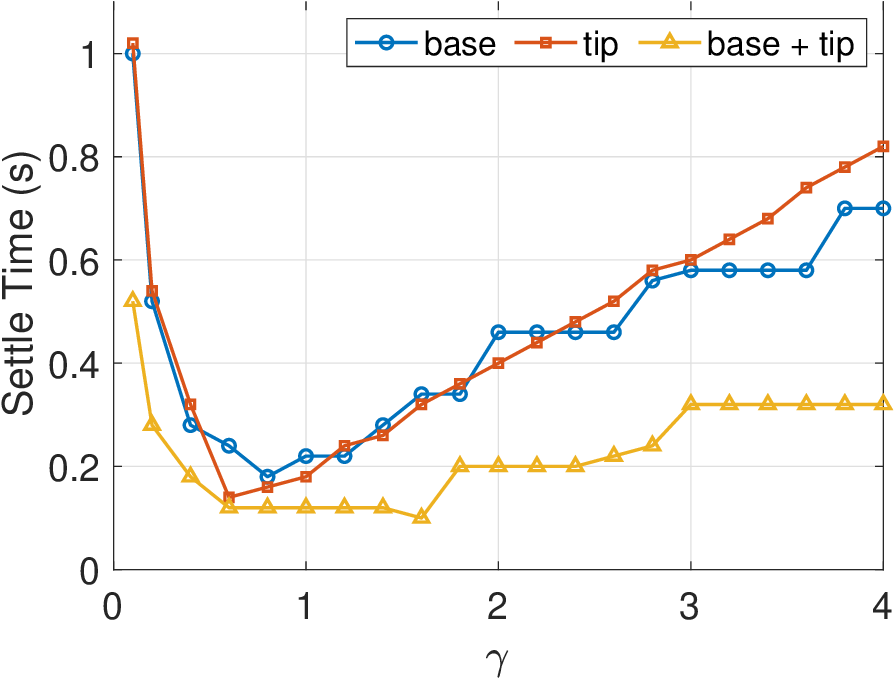}
    \end{subfigure}

    \caption{\textbf{Left}: Initial configurations of the ground-truth and estimated rods. The estimate is deliberately initialized with a perturbed configuration to evaluate convergence. \textbf{Middle}: Ground-truth and estimated tip-position trajectories using the base observer with $\gamma=1$. \textbf{Right}: Settling times of the observers as functions of the gain scaling factor $\gamma$. As $\gamma$ increases, the settling times initially decrease, reach a minimum near $\gamma = 1$, and then increase.}
    \label{fig:sim balanced}
\end{figure*}

We conducted simulations to validate the reference optimal gains. To simulate a ``balanced'' Cosserat rod, we chose
\begin{align*}
    \M = 10I_{6\times6}, \quad \K = \SI{1e4}I_{6\times6},
\end{align*}
where $\M$ and $\K$ are mass and stiffness matrices, respectively. These values do not correspond to any specific physical robot but were intentionally chosen to provide an idealized setting for quantitatively testing the predictions based on the linearized model.

We first simulated a dynamic motion of the rod in a hanging configuration. The rod was released from a holding force at the tip, allowing it to oscillate freely. The resulting trajectories were saved as ground truth. Then, the observer algorithm was initialized from a straight configuration. See Fig.~\ref{fig:sim balanced} left. In the estimation, as the rod did not receive any external inputs except gravity, any motion was entirely driven by the corrective boundary conditions. We tested the base and tip observers separately, and their combination. The P term was disabled in this study.

The theoretical observer gains $\Gamma_0'$ and $\Gamma_1'$ were computed according to \eqref{eq:ref gain diagonal}. To assess performance across varying gain magnitudes, we conducted multiple tests with gains scaled as
\begin{align}
    \Gamma_0 = \gamma\Gamma_0', \quad \Gamma_1 = \gamma\Gamma_1',
\end{align}
where $\gamma$ is varied from 0.2 to 4. For each test, we computed the settling time of the tip position, defined as the time when the estimation error fell below $2\%$ of the initial error.

\medskip
\noindent\textbf{Results.} The results are summarized in Fig.~\ref{fig:sim balanced}. The middle panel shows that the estimated tip-position trajectory obtained using the base observer ($\gamma=1$) converges to the ground truth within 0.2 seconds, whereas the right panel shows the relationship between settling time and the gain scaling factor $\gamma$. As $\gamma$ increases from small values, the settling time initially decreases, reaches a minimum near $\gamma = 1$, and then increases. This trend is consistent with the theoretical relationship between convergence rate and observer gain shown in Fig.~\ref{fig:mu-max}. All observers achieve their fastest convergence near $\gamma = 1$, indicating that the reference optimal gains provide a quantitatively accurate prediction even though they are obtained using a linear approximation. Moreover, the combined observer consistently outperforms the individual observers, confirming that dual dissipation accelerates convergence.



\section{Experimental Validation}
\label{sec:experiment}

In this section, we describe the design of a physical continuum robot and the experimental procedures used to evaluate the performance of the boundary observers and the impact of observer gains.

\subsection{Hardware Setup}
\label{sec:experimental setup}

The robot consists of a spring steel rod (McMaster-Carr, Illinois, USA) as the backbone, with nine equally spaced disks attached. The nominal properties of the rod are listed in Table~\ref{tab:robot parameter}. A 6-axis FT sensor (MINI40 SI-40-2, ATI Industrial Automation, North Carolina, USA) was mounted at the robot's base to measure its real-time internal wrenches.
Two parallel tendons were used to actuate the robot. Tendon 1 was attached to the last disk at the tip, so pulling it bent the entire robot. Tendon 2 was attached to the disk at the midpoint, so pulling it bends only the first half of the robot. Each tendon's base end was tied to a force gauge (M3-20, Mark-10, New York, USA) to measure its real-time tension.
To establish a ground truth model of the robot's state, markers were placed on five equally spaced disks and tracked using a motion capture system (Vicon, Oxford, UK). All measurement data were recorded at \SI{100}{\hertz} on a PC for offline validation. 
When feeding the recorded data into our algorithm, which operated at a different rate, linear interpolation was used as needed. The complete experimental setup is depicted in Fig.~\ref{fig:exp tendon driven snapshot}~(a).

\begin{table}[!htbp]
\small
\caption{Properties of the spring steel rod.}
\setlength{\tabcolsep}{0.5em}
\renewcommand{\arraystretch}{1.2}
    \centering
    \begin{tabular}{rl}
        \hline
        Length $L$ & \SI{600}{\milli\meter} \\
        Radius $r$ & \SI{0.8}{\milli\meter} \\
        Density $\rho$ & \SI{7.87e3}{\kg\per\m\cubed} \\
        Young's modulus $E$ & \SI{200}{\giga\pascal} \\
        Shear modulus $G$ & \SI{76.92}{\giga\pascal} \\
        \hline
    \end{tabular}
    \label{tab:robot parameter}
\end{table}
\subsection{Algorithm Implementation}

The model parameters $\M(s)$ and $\K(s)$ were computed from the values in Table~\ref{tab:robot parameter}. The manufacturer provided the nominal length, radius, and density. However, the density was calibrated to \SI{4.48e4}{\kg\per\meter\cubed} to account for the additional mass of the spacer disks and markers, which was assumed to be uniformly distributed along the rod. The robot was then installed horizontally, and different weights were applied at the tip to determine the Young's modulus experimentally, following a procedure commonly used in prior work~\citep{rucker2011statics, feliu2024dynamic}. The inertia and stiffness matrices were then computed as $\M = \diag(\M_a, \M_l)$ and $\K = \diag(\K_a, \K_l)$, where
\begin{align*}
    \M_a & =\diag(1,1,2)\rho\pi r^4/4 \\
    & =\diag(1.44 \times 10^{-8},\ 1.44 \times 10^{-8},\ 2.88 \times 10^{-8}), \\
    \M_l & =I_{3\times3}\rho\pi r^2 \\
    & =\diag(9.00 \times 10^{-2},\ 9.00 \times 10^{-2},\ 9.00 \times 10^{-2}), \\
    \K_a & =\diag(E,E,2G)\pi r^4/4 \\
    & =\diag(6.43 \times 10^{-2},\ 6.43 \times 10^{-2},\ 4.84 \times 10^{-2}), \\
    \K_l & =\diag(G,G,E)\pi r^2 \\
    & =\diag(1.51 \times 10^{5},\ 1.51 \times 10^{5},\ 4.02 \times 10^{5}). 
\end{align*}
We observe that the angular and linear components of $\K$ differ in magnitude by approximately $10^6$ to $10^7$. In this case, the linear strain remains nearly constant, meaning that the robot behaves as a Kirchhoff rod. As discussed previously, the linear states can be fully determined by the angular states and vice versa in the planar case. Additionally, the theoretically derived optimal gains based on a linear approximation may deviate from the actual optimal values because of the system's nonlinearities. To investigate observer performance and the effect of gain selection, we conducted three sets of experiments. In all experiments, the observer algorithms were implemented numerically in MATLAB using the shooting method described in Algorithm~\ref{alg:boundary observer}. The arc-length domain was discretized into 30 points, and the temporal resolution was set to \SI{30}{\hertz}.

\begin{figure}[t]
    \centering
    \begin{subfigure}[b]{0.2\textwidth}
        \centering
        \includegraphics[height=4.4cm]{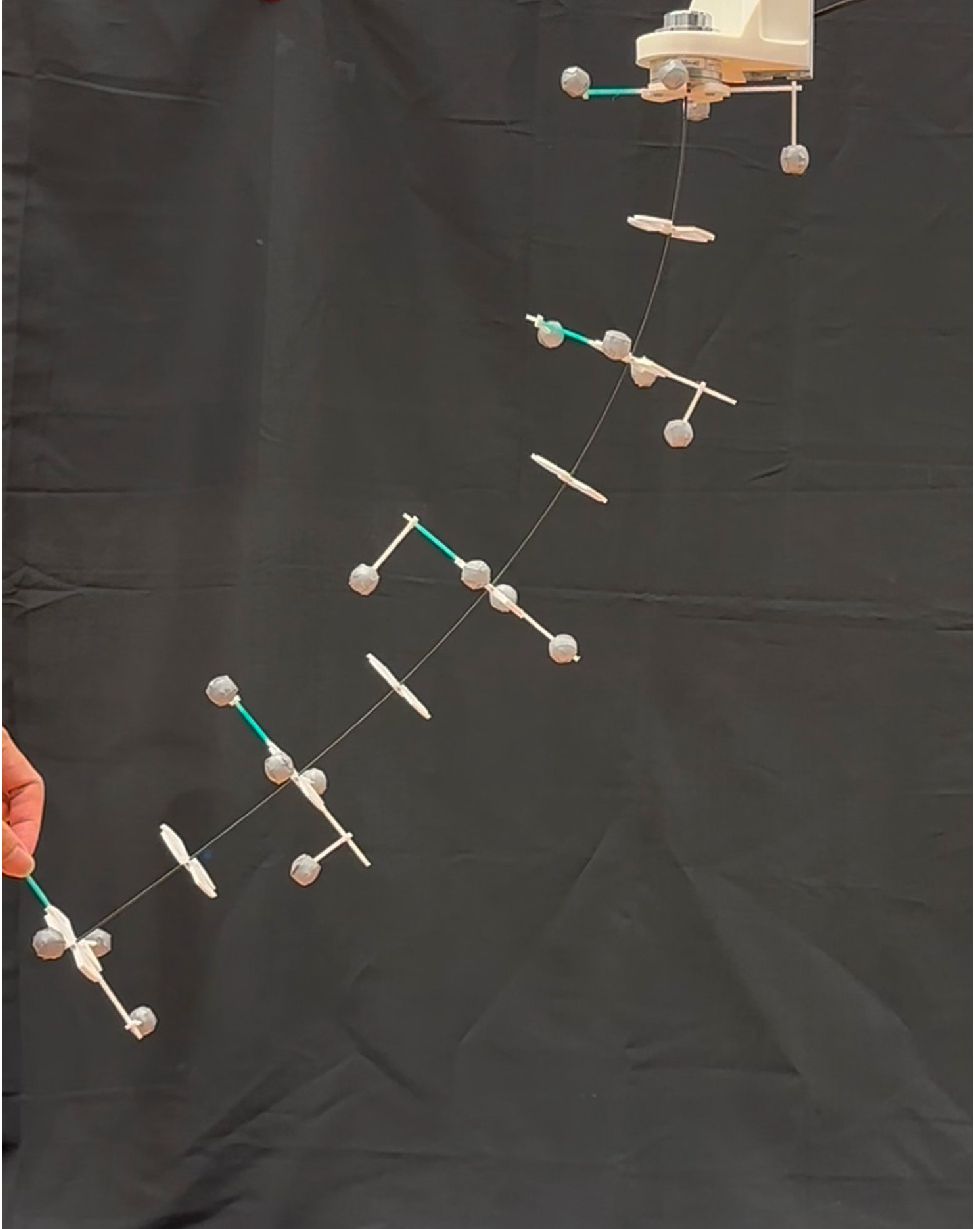}
    \end{subfigure}
    \hfill
    \begin{subfigure}[b]{0.28\textwidth}
        \centering
        \includegraphics[height=4.4cm]{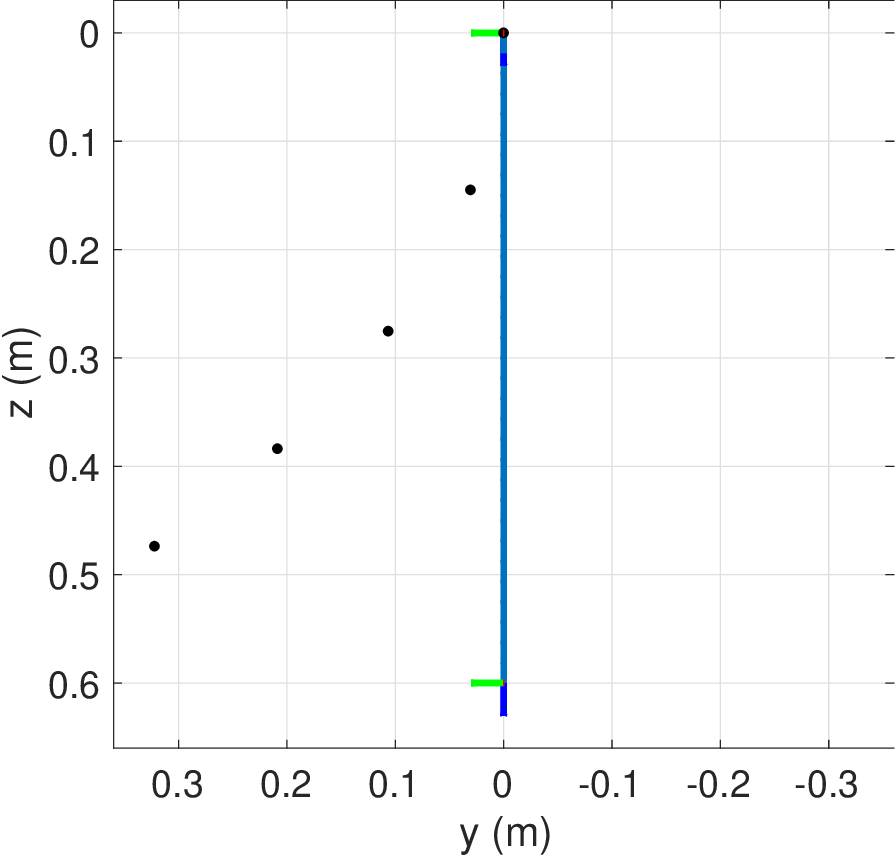}
    \end{subfigure}

    \caption{\textbf{Left}: The rod is released to induce free oscillations. \textbf{Right}: Initial configurations of the ground truth and the estimated rod. The estimate is deliberately initialized with a deviated configuration to evaluate its convergence behavior.}
    \label{fig:exp oscillation initial}
\end{figure}

\begin{figure*}[htbp]
    \centering
    \begin{subfigure}[b]{0.33\textwidth}
        \centering
        \includegraphics[width=1\textwidth]{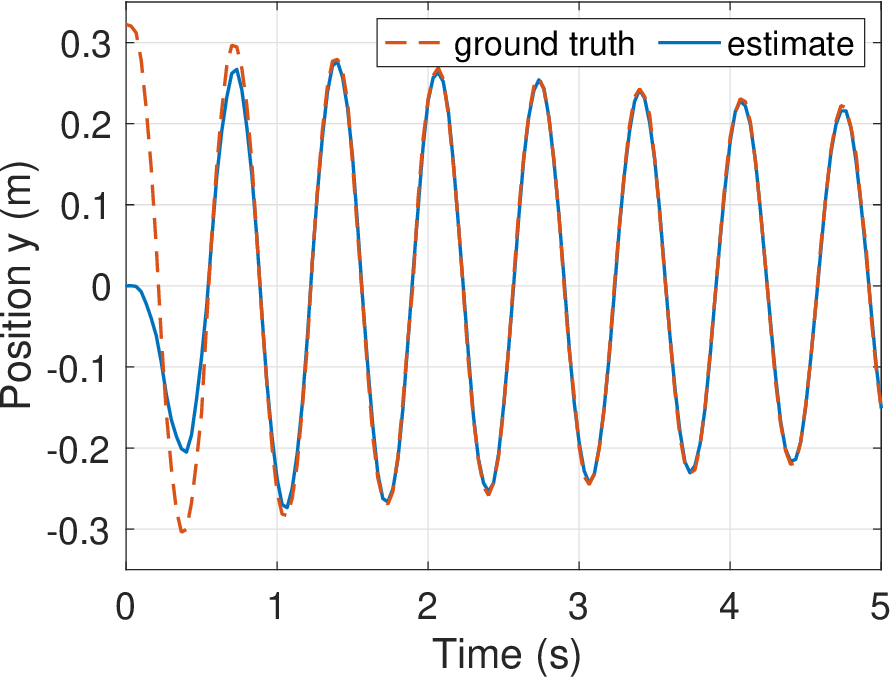}
    \end{subfigure}
    \hfill
    \begin{subfigure}[b]{0.33\textwidth}
        \centering
        \includegraphics[width=1\textwidth]{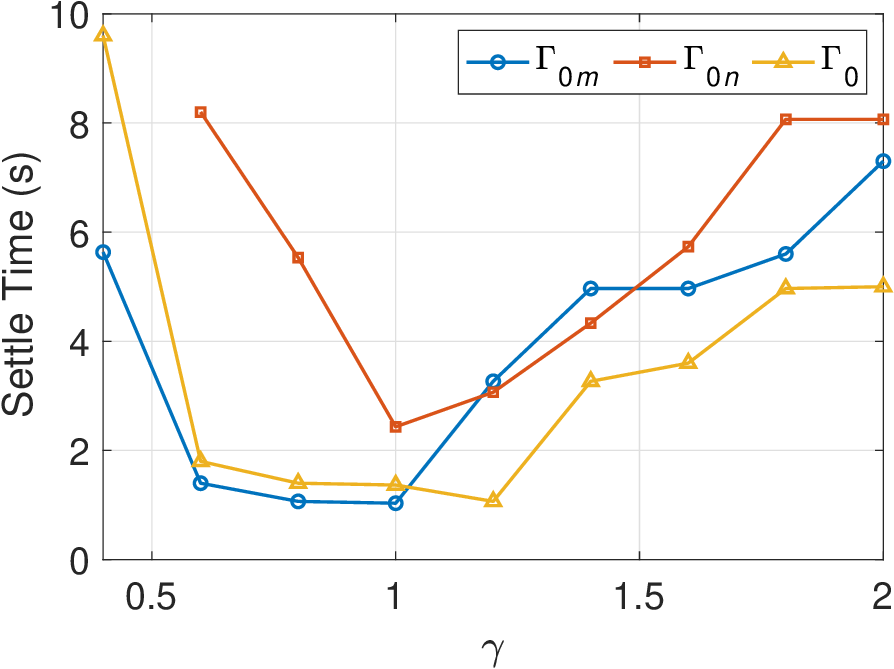}
    \end{subfigure}
    \hfill
    \begin{subfigure}[b]{0.33\textwidth}
        \centering
        \includegraphics[width=1\textwidth]{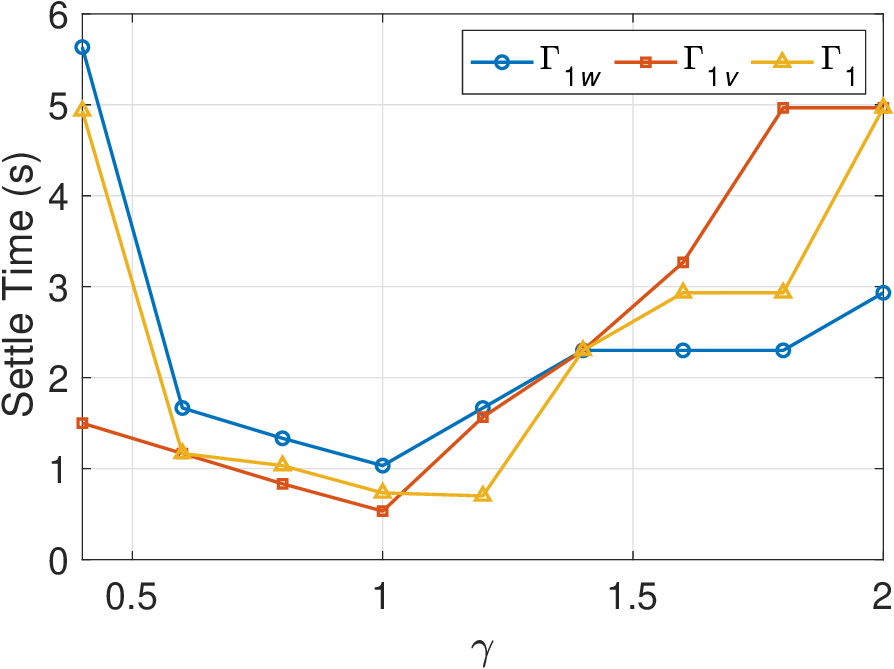}
    \end{subfigure}

    \caption{\textbf{Left}: Ground-truth and estimated tip-position trajectories using the base observer with $\gamma=1$. \textbf{Middle}: Settling times of base-observer variants as functions of the gain scaling factor $\gamma$. \textbf{Right}: Settling times of tip-observer variants as functions of the gain scaling factor $\gamma$.
    }
    \label{fig:exp oscillation settle time}
\end{figure*}

\begin{figure*}[!htbp]
    \centering
    \includegraphics[width=0.99\textwidth]{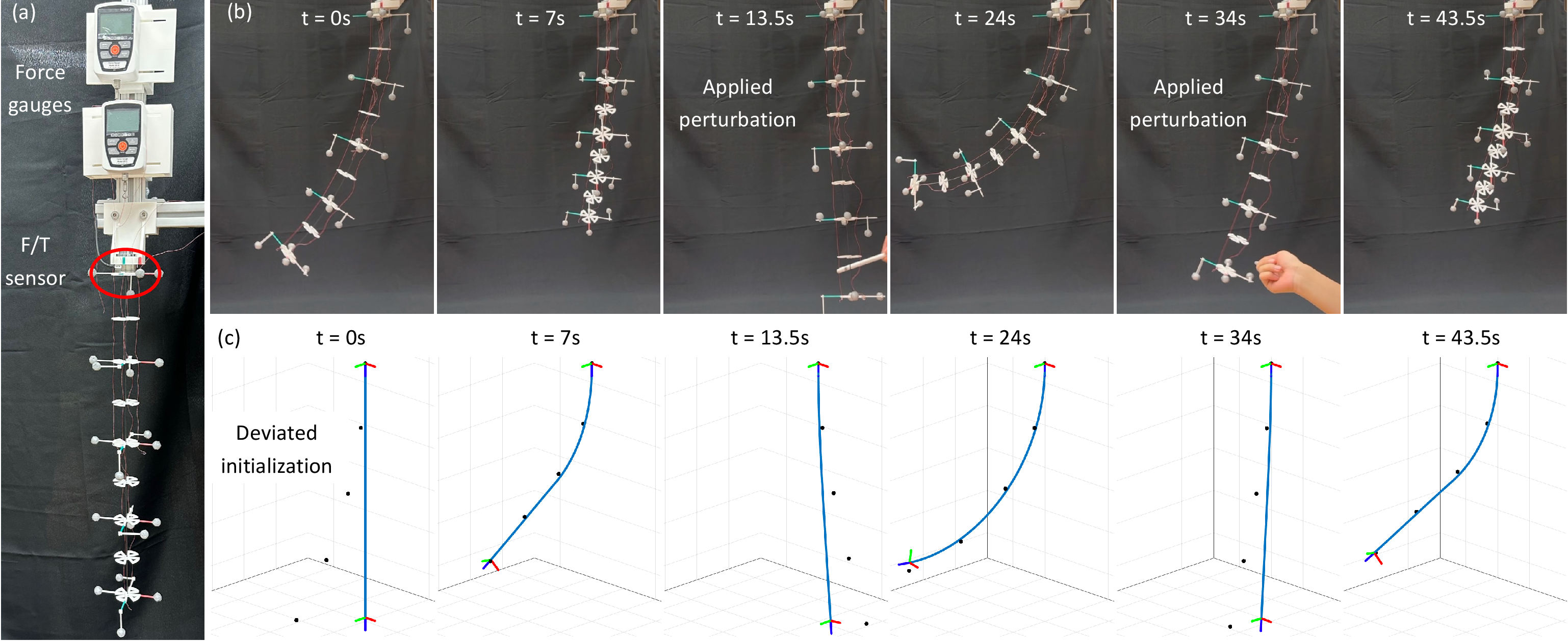}
    \caption{(a) Experimental Setup: The robot is actuated by two parallel tendons. Tracking markers are placed on five spacer disks to measure their position and orientation, serving as ground truth. An FT sensor is mounted at the robot's base to measure its real-time internal wrenches, while a force gauge is attached to each tendon to record its real-time tension.
    (b) Experimental Snapshots: The two tendons are pulled to generate spatial motions. At near $t = 13.5$ and $34$ seconds, external perturbations are applied to the robot to induce high-frequency free vibrations.
    (c) Estimated Configuration for Corresponding Snapshots: The black dots represent the ground truth positions of the robot, while the blue curve denotes the estimated configuration using the base observer. 
    At $t=0$, the algorithm is deliberately initialized when the robot is already in motion.
    }
    \label{fig:exp tendon driven snapshot}
\end{figure*}

\subsection{Experiment 1: Free Oscillation}

This experiment is similar to the simulation study. The rod was installed vertically. It was initially held at the tip and then released, allowing it to oscillate freely; see Fig.~\ref{fig:exp oscillation initial} (left). Then, the observer algorithm was initialized from a straight configuration; see Fig.~\ref{fig:exp oscillation initial} (right). Due to this incorrect initialization, the estimated states would have remained static in the absence of boundary corrections. This experiment corresponds to a planar Kirchhoff case, in which the linear and angular components are fully determined by each other. To evaluate whether the observer converges when provided with only linear or only angular state measurements, we injected individual measurements, including base moment, base force, tip angular velocity, and tip linear velocity, into the observer separately. The corresponding optimal gains were determined using the bracketing search algorithm outlined in Algorithm~\ref{alg:gain tuning} and were found to be:
\begin{align*}
    \Gamma_{0m}^{*}  &=0.00035\Gamma_{0m}',  &\quad \Gamma_{0n}^{*}  &=100\Gamma_{0n}', \\
    \Gamma_{1w}^{*}  &=450\Gamma_{1w}',   &\quad \Gamma_{1v}^{*}  &= 0.0008\Gamma_{1v}'.
\end{align*}
To account for the fact that the linear and angular components carry equivalent information, we set
\begin{align}   \label{eq:Gamma01=average}
    \Gamma_0^{*} = (\Gamma_{0m}^{*} + \Gamma_{0n}^{*})/2, \quad
    \Gamma_1^{*} = (\Gamma_{1w}^{*}  + \Gamma_{1v}^{*})/2,
\end{align}
and ran a series of tests for the six cases separately with gains scaled as
\begin{align*}
    \Gamma_{0m}=\gamma\Gamma_{0m}^{*}, \quad \Gamma_{0n}=\gamma\Gamma_{0n}^{*}, \quad \Gamma_0=\gamma\Gamma_0^{*}, \\
    \Gamma_{1w}=\gamma\Gamma_{1w}^{*}, \quad \Gamma_{1v}=\gamma\Gamma_{1v}^{*}, \quad \Gamma_1=\gamma\Gamma_1^{*}.
\end{align*}
Only one of the above gains was enabled at a time; the others were set to zero. Because of model uncertainty, it was difficult to reduce the estimation error below 2\% of the initial error. Therefore, for all physical experiments, we defined the settling time as the time at which the estimation error dropped below 5\% of the robot's total length.

\medskip
\noindent\textbf{Results.} The results are summarized in Fig.~\ref{fig:exp oscillation settle time}. The left panel shows that the estimated tip $y$-position trajectory obtained using the base observer ($\gamma = 1$) converges to the ground truth within \SI{1.5}{\second}. The middle panel shows the relationship between settling time and the gain scaling factor $\gamma$ for $\Gamma_{0m}$, $\Gamma_{0n}$, and $\Gamma_0$ in the base observer. The right panel presents corresponding results for the tip observer, showing the dependence of settling time on $\Gamma_{1w}$, $\Gamma_{1v}$, and $\Gamma_1$. These results lead to three important observations. First, they confirm that, in the Kirchhoff case, convergence can be achieved using only angular measurements. Consequently, the tip observer can be implemented using only embedded sensors such as IMUs. Second, the qualitative ``decrease-then-increase'' trend in settling time with respect to $\gamma$ remains valid, although the actual optimal value differs from the theoretical reference value given in~\eqref{eq:ref gain}. Third, $\Gamma_0$ and $\Gamma_1$ achieve the fastest convergence near $\gamma = 1$, even though their values are predicted directly from~\eqref{eq:Gamma01=average} rather than tuned through a search. This observation further supports the conclusion that linear and angular measurements carry equivalent information in the Kirchhoff case.

\subsection{Experiment 2: Tendon-Driven Motion}

In this experiment, we manually operated the two force gauges to pull the tendons and generate spatial motions of the robot. To evaluate observer robustness, we occasionally applied external perturbations by stirring the robot's tip using a stick to induce free vibrations. A video demonstration is provided in the accompanying multimedia submission. These perturbations were introduced to evaluate the observers' ability to recover from unknown external inputs. Due to the distributed approximation of the actuation wrench in the tendon model and the omission of tendon friction, this experiment involved significantly higher modeling uncertainty than Experiment~1. We tested four observer configurations: base, tip D, tip PD, and the combined observer (base + tip PD). The observer gains were set to the actual optimal values identified in Experiment~1. For the observers involving proportional feedback, the proportional gain was set to $\Gamma_\mathrm{P} = 20 \Gamma_\mathrm{D}$. All observer algorithms were initialized when the robot began moving to induce a large initial estimation error. A sequence of experimental snapshots is shown in Fig.~\ref{fig:exp tendon driven snapshot}~(b).

\begin{figure*}[!htbp]
    \centering
    \includegraphics[width=0.98\textwidth]{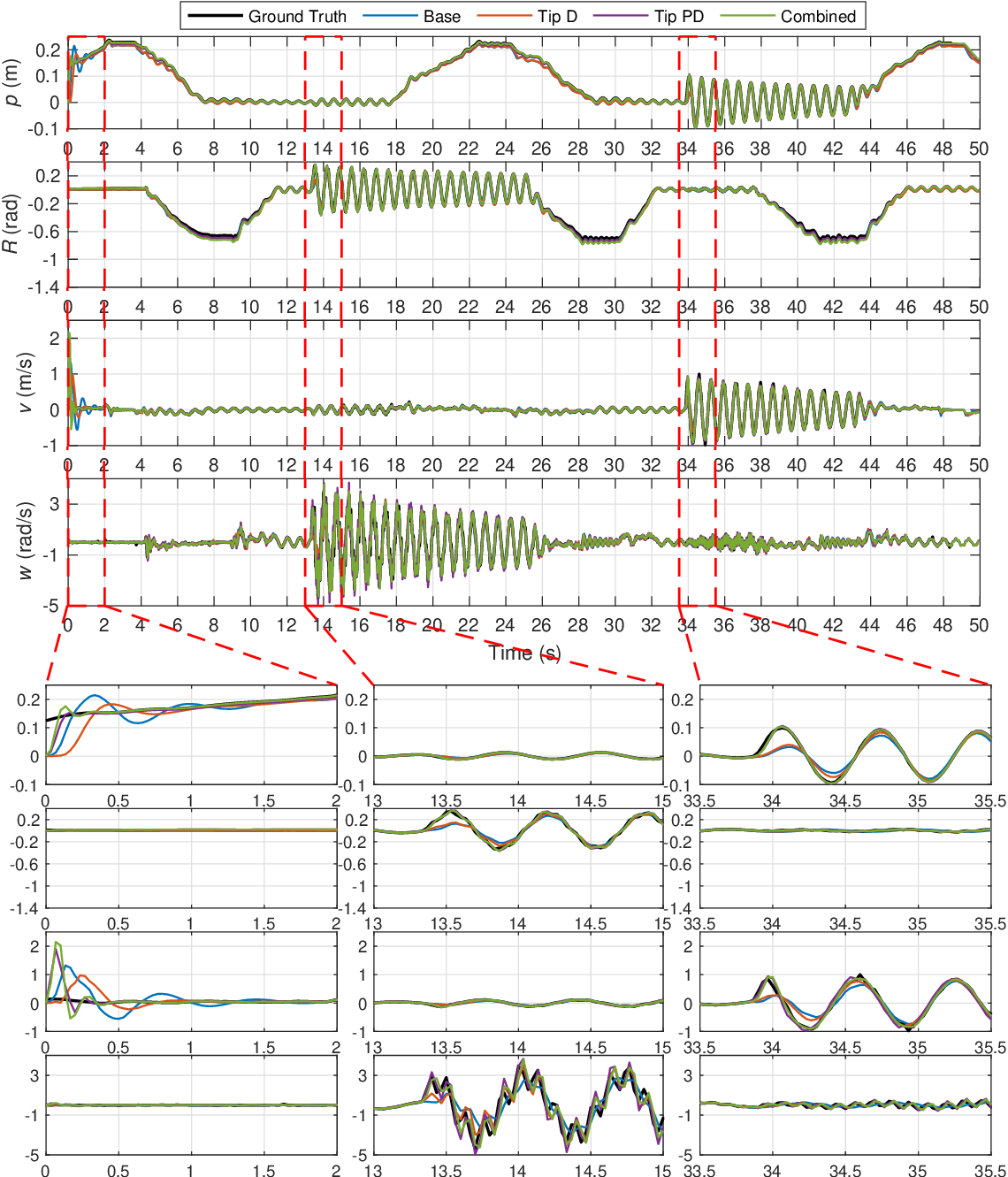}
    \caption{Comparison of the ground-truth and estimated position and orientation trajectories (expressed as XYZ Euler angles) at the three-quarter point of the robot. Only the $y$-components are plotted. The estimation algorithms are initialized while the robot is already in motion. All observers converge to the ground truth within 2 seconds. The tip PD and combined observers exhibit smaller steady-state errors. The base and tip D observers exhibit temporary deviations near $t = 13.5$ and $34$ seconds because of unknown perturbations but reconverge within 1 second. The tip PD and combined observers continue to track the ground truth in the presence of unknown perturbations. The combined observer converges fastest and exhibits the least oscillation.}
    \label{fig:exp tendon driven tracking}
\end{figure*}

\begin{figure*}
    \centering
    \includegraphics[width=0.99\textwidth]{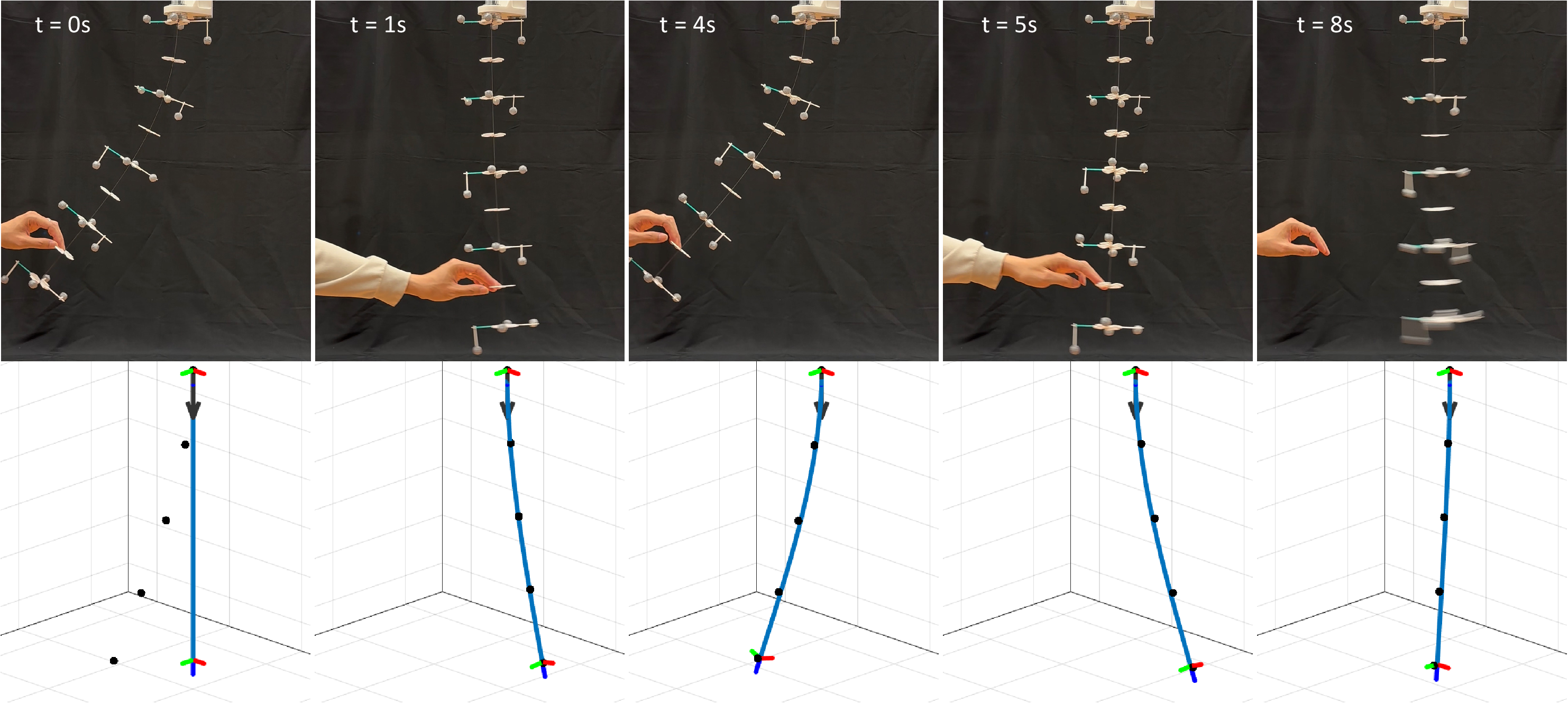}
    \caption{\textbf{Top}: Experimental Snapshots. \textbf{Bottom}: Estimated Configurations. The black dots represent the ground truth positions, while the blue curve denotes the estimated position using the tip PD observer.}
    \label{fig:exp unknown drag snapshot}
\end{figure*}

\begin{figure*}[htbp]
    \centering
    \begin{subfigure}[b]{0.33\textwidth}
        \centering
        \includegraphics[width=1\textwidth]{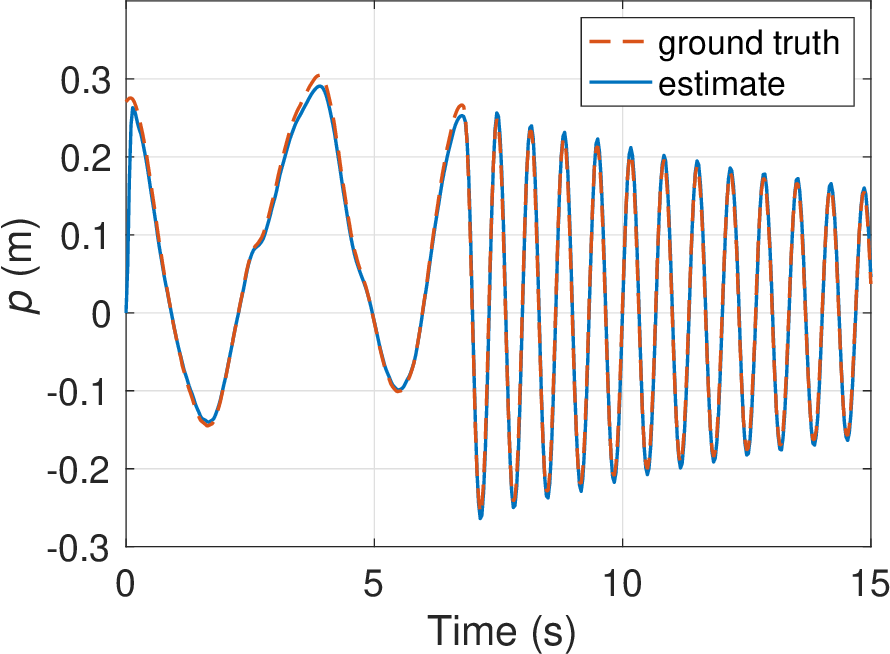}
    \end{subfigure}
    \hfill
    \begin{subfigure}[b]{0.33\textwidth}
        \centering
        \includegraphics[width=1\textwidth]{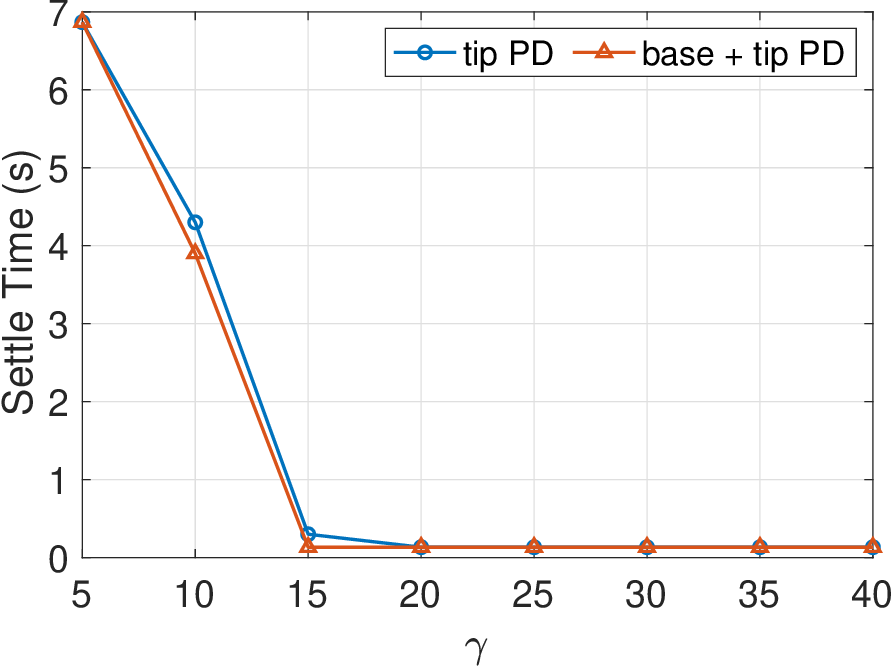}
    \end{subfigure}
    \hfill
    \begin{subfigure}[b]{0.33\textwidth}
        \centering
        \includegraphics[width=1\textwidth]{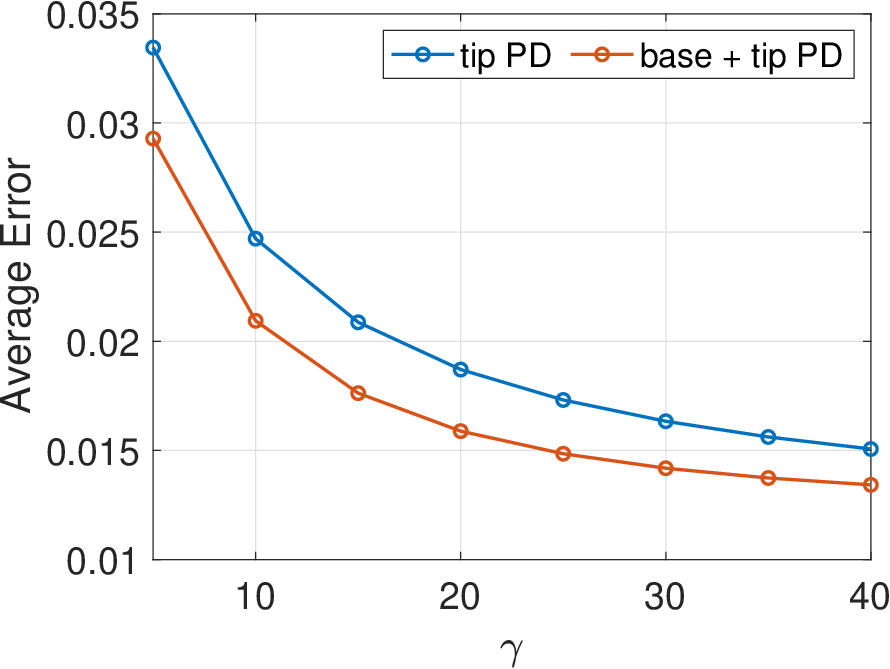}
    \end{subfigure}

    \caption{\textbf{Left}: Ground truth and estimated tip position-$y$ trajectories using the tip PD observer with $\gamma = 20$. \textbf{Middle}: Settling times of the tip PD and combined observers as functions of the gain scaling factor $\gamma$. \textbf{Right}: Average tip position errors of the tip PD and combined observers as functions of $\gamma$.}
    \label{fig:exp unknown drag settle time}
\end{figure*}

\medskip
\noindent\textbf{Results.}
In Fig.~\ref{fig:exp tendon driven snapshot}, we present the recorded ground truth positions and the estimated configurations using the base observer across six experimental snapshots. The ground truth and estimated trajectories for position, orientation (expressed as XYZ Euler angles), and linear and angular velocities (in the local frame) at the three-quarter point of the robot are shown in Fig.~\ref{fig:exp tendon driven tracking}. Only the $y$-component is shown, as the other components exhibit similar trends. 

For position and orientation estimation, all observers converge within \SI{2}{\second}, although the steady-state errors vary. The tip PD and combined observers exhibit smaller steady-state errors due to the proportional correction term. Similar trends are observed in the linear and angular velocity estimates. The transient phase following initialization lasted at most \SI{2}{\second}. Velocity overshoots are primarily caused by large initialization errors and could be mitigated by initializing the observer closer to the true robot state. The base and tip D observers exhibit temporary deviations at $t = 13.5$ and \SI{34}{\second} due to unknown perturbations, but re-converge within \SI{1}{\second} and successfully capture high-frequency vibrations. In contrast, the tip PD and combined observers maintain accurate tracking performance even under such perturbations. The average estimation errors are reported in Table~\ref{tab:avg error}. The tip PD and combined observers demonstrate comparable performance and generally achieve lower errors compared to the base and tip D observers. These results suggest that the inclusion of proportional correction enhances robustness to perturbations and reduces steady-state errors.

During large-angle deformations, the estimated trajectories by the base and tip D observers show larger deviations from the ground truth. Two primary factors may contribute to this discrepancy. First, under large deformations, the assumed linear constitutive law~\eqref{eq:linear constitutive law} becomes less accurate, and increased friction between the tendons and spacer disks introduces additional modeling errors. Second, the assumption of uniform density and diagonal mass and stiffness matrices $\M$ and $\K$ may fail to capture the physical asymmetries of the robot. This issue can be mitigated through system identification to find nonuniform $\M(s)$ and $\K(s)$ before performing state estimation.

\begin{table}[!htbp]
\setlength{\tabcolsep}{3pt}
\centering
\caption{Average estimation errors for position $p$, rotation $R$, linear velocity $v$, and angular velocity $w$ by the four observers. The position errors are represented as percentages of the robot's total length.}
\label{tab:avg error}
\begin{tabular}{l|cccc}
\hline
 & $p$ (\%) & $R$ (\si{\degree})
 & $v$ (\si{\meter/\second})
 & $w$ (\si{\radian/\second}) \\
\hline
Base & 1.9  &  2.7  &  0.064  &  0.39 \\
Tip D & 1.7  &  2.5  &  0.052  &  0.31 \\
Tip PD & 0.9  &  1.4  &  0.047  &  0.28 \\
Comb & 1.0   & 2.1 &   0.029  &  0.28 \\
\hline
\end{tabular}
\end{table}

\begin{table}[!htbp]
\setlength{\tabcolsep}{3pt}
\centering
\caption{Real-time factor of the four observers for varying spatial and temporal discretizations.}
\label{tab:computation time}
\begin{tabular}{l|cccc}
\hline
 & 30 pts & 35 pts & 40 pts & 45 pts \\
\hline
\SI{30}{\hertz} Base   & 1.52 & 1.31 & 1.14 & 1.00 \\
\SI{30}{\hertz} Tip-D  & 1.52 & 1.34 & 1.18 & 1.05 \\
\SI{30}{\hertz} Tip-PD & 1.51 & 1.31 & 1.15 & 1.02 \\
\SI{30}{\hertz} Comb   & 1.52 & 1.30 & 1.14 & 1.00 \\
\hline
\SI{35}{\hertz} Base   & 1.27 & 1.11 & 0.95 & 0.87 \\
\SI{35}{\hertz} Tip-D  & 1.27 & 1.13 & 0.99 & 0.86 \\
\SI{35}{\hertz} Tip-PD & 1.24 & 1.10 & 0.95 & 0.86 \\
\SI{35}{\hertz} Comb   & 1.28 & 1.11 & 0.94 & 0.87 \\
\hline
\SI{40}{\hertz} Base   & 1.10 & 0.95 & 0.81 & 0.74 \\
\SI{40}{\hertz} Tip-D  & 1.13 & 0.97 & 0.86 & 0.76 \\
\SI{40}{\hertz} Tip-PD & 1.10 & 0.95 & 0.83 & 0.74 \\
\SI{40}{\hertz} Comb   & 1.10 & 0.95 & 0.83 & 0.75 \\
\hline
\end{tabular}
\end{table}

Table~\ref{tab:computation time} presents the real-time factor, defined as the ratio between the simulated time and the actual wall-clock time required to perform the computation (with values $\geq 1$ indicating real-time capability), under varying levels of spatial and temporal discretization. All computations are conducted on a 64-bit Windows workstation equipped with a 13th Gen Intel\textsuperscript{\textregistered} Core\texttrademark{} i9-13900 processor at \SI{2}{\giga\hertz} and \SI{64}{\giga\byte} of RAM. As expected, the real-time factors decrease approximately linearly with increasing resolution and remain comparable across different observer configurations, indicating that the inclusion of additional measurement inputs does not significantly affect computational efficiency. These results are obtained using a MATLAB implementation without real-time optimization. As noted earlier, an optimized C++ implementation can achieve substantially higher computational efficiency \citep{till2019real}.

\subsection{Experiment 3: Motion Caused by Unknown Inputs}

This experiment was conducted to validate the robustness of the proportional term when the robot was subjected to completely unknown inputs. The robot was manually moved through space and then released after a few seconds to oscillate freely. The actuation was entirely unknown to the estimation algorithm; without boundary correction, the estimated robot would therefore have remained static. We tested the tip PD and combined observers in this scenario. As before, the estimation algorithm was initialized while the robot was already in motion to examine its convergence behavior. Experimental snapshots and the corresponding estimated configurations are shown in Fig.~\ref{fig:exp unknown drag snapshot}. A series of tests was performed with $\Gamma_0 = \Gamma_0^{*}$, while the PD gains were scaled as
\begin{align*}
    \Gamma_{\mathrm{D}} = \gamma\Gamma_1^{*}, \quad \Gamma_{\mathrm{P}} = 20\gamma\Gamma_1^{*}.
\end{align*}
Maintaining a fixed ratio between the P and D terms is important to avoid excessive overshoot under large $\Gamma_{\mathrm{P}}$ values.

\medskip
\noindent\textbf{Results.} The results are summarized in Fig.~\ref{fig:exp unknown drag settle time}. The left panel shows that the estimated tip $y$-position trajectory obtained using the tip PD observer with $\gamma = 20$ converges to the ground truth within $0.2$ seconds. The middle panel illustrates the relationship between settling time and the gain scaling factor $\gamma$. As $\gamma$ increases, the settling time initially decreases and then plateaus, possibly because the convergence rate has reached its characteristic limit. The right panel shows the relationship between average position error and $\gamma$: larger values of $\gamma$ consistently yield smaller average errors. This result is consistent with the intuition that a larger proportional term helps reduce steady-state error. Notably, the combined observer achieved lower average errors across all values of $\gamma$ because of its dual dissipative boundary conditions. Beyond a certain range, further increasing the gain scaling factor does not noticeably improve settling time or steady-state accuracy. However, doing so may cause numerical instability by injecting excessively large correction wrenches when the tip error is large.

\begin{figure*}[htbp]
    \centering
    \begin{subfigure}[b]{0.49\textwidth}
        \centering
        \includegraphics[width=\textwidth]{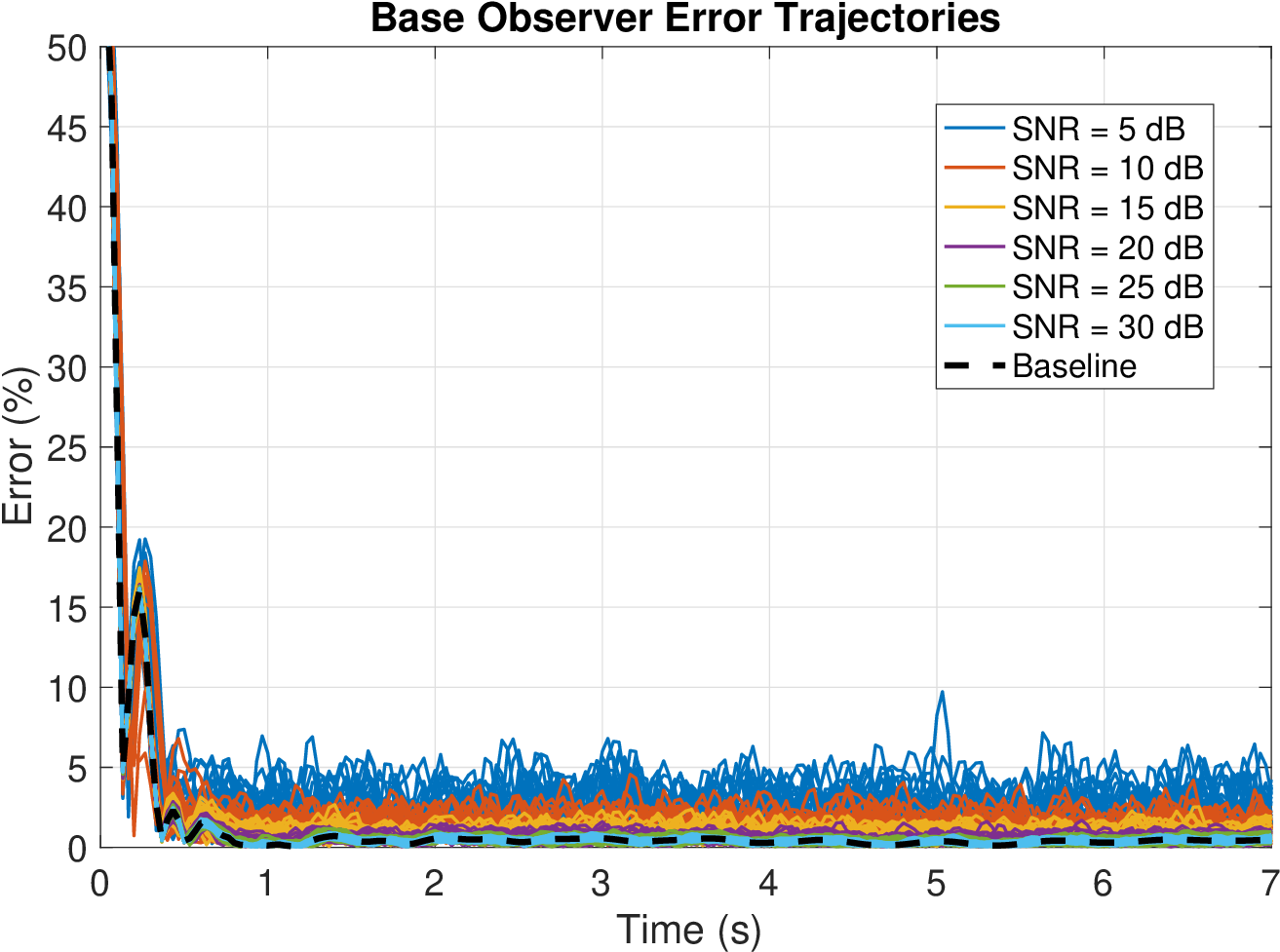}
    \end{subfigure}
    \hfill
    \begin{subfigure}[b]{0.49\textwidth}
        \centering
        \includegraphics[width=\textwidth]{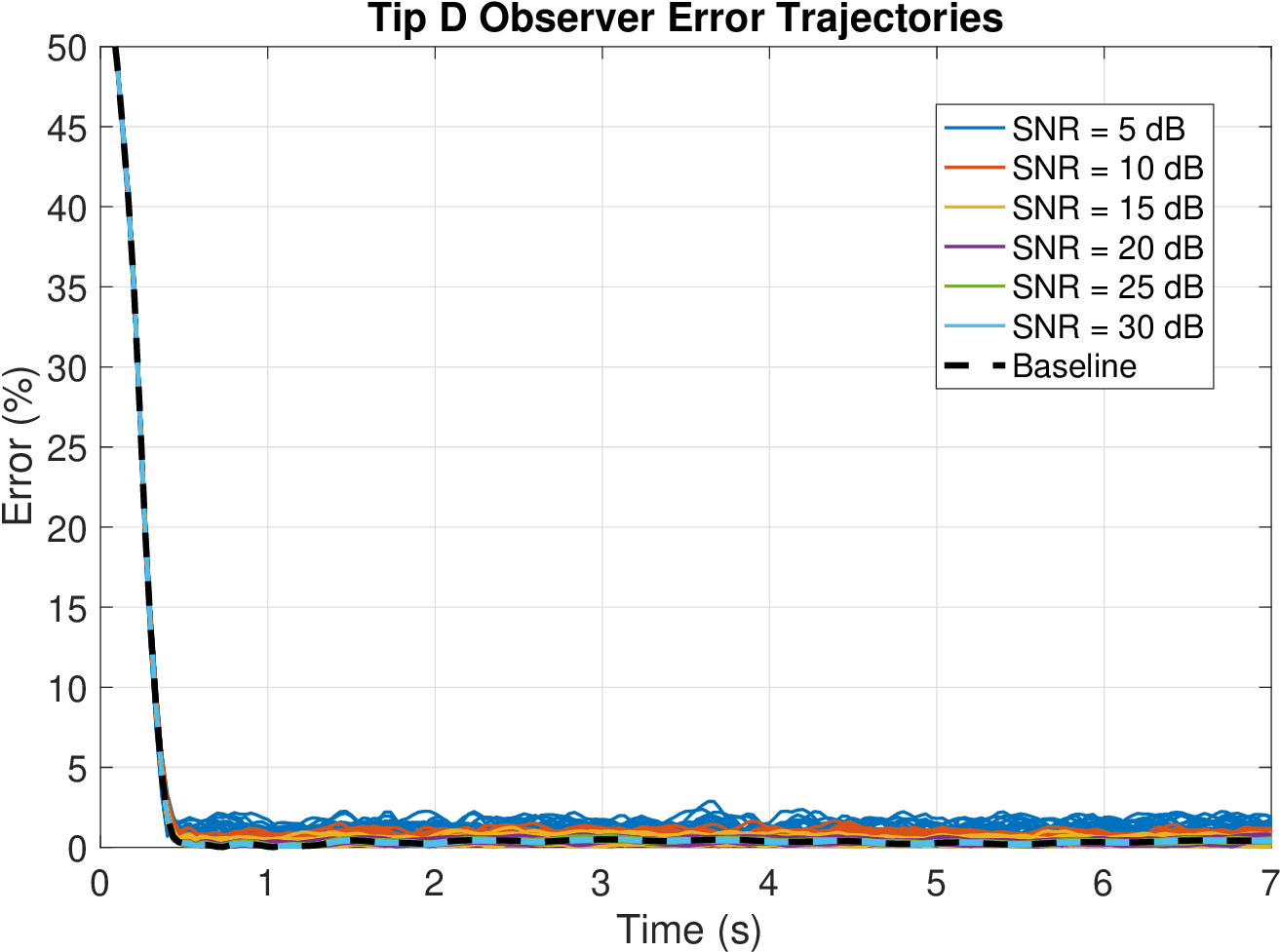}
    \end{subfigure}

    \begin{subfigure}[b]{0.49\textwidth}
        \centering
        \includegraphics[width=\textwidth]{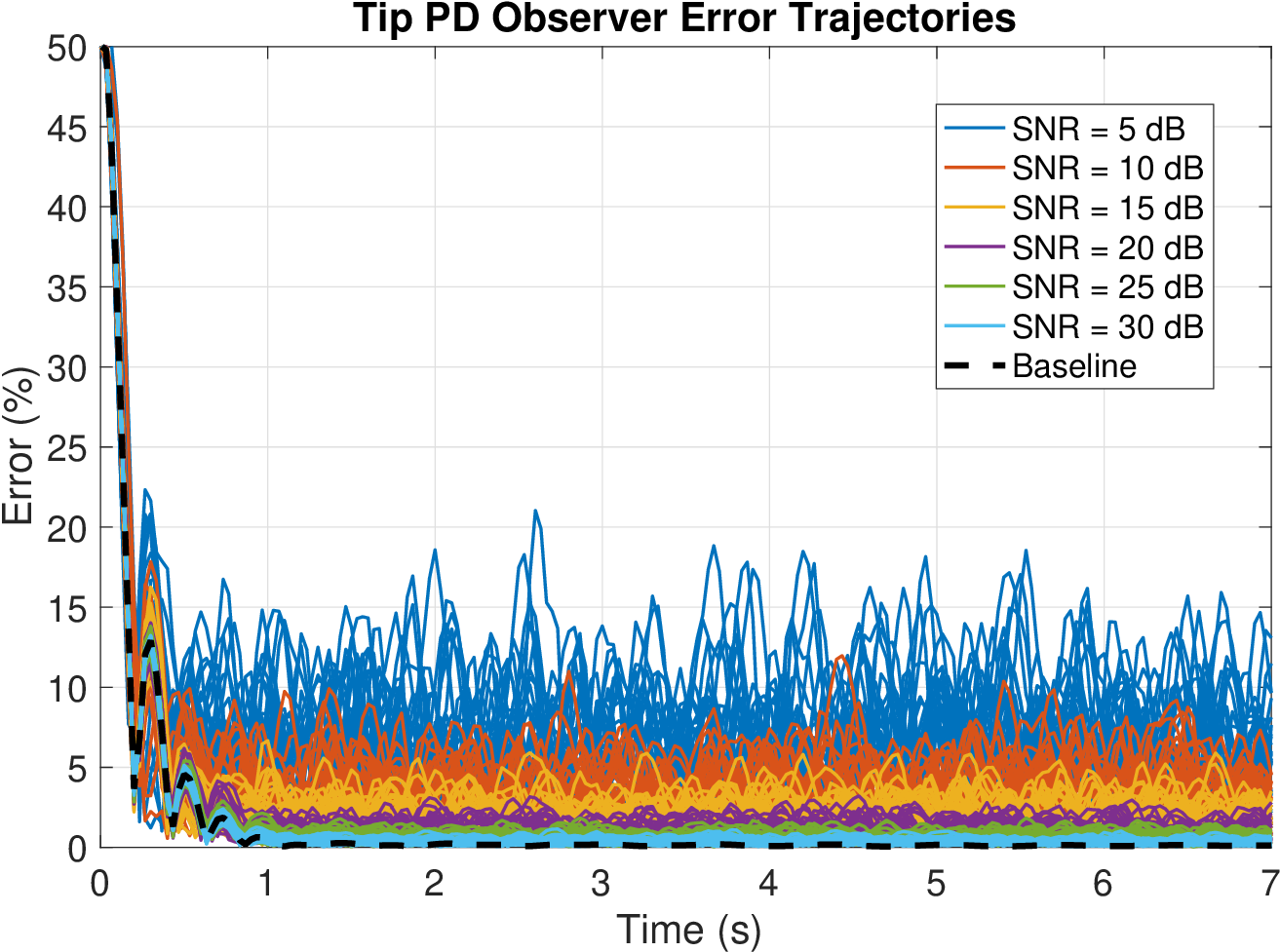}
    \end{subfigure}
    \hfill
    \begin{subfigure}[b]{0.49\textwidth}
        \centering
        \includegraphics[width=\textwidth]{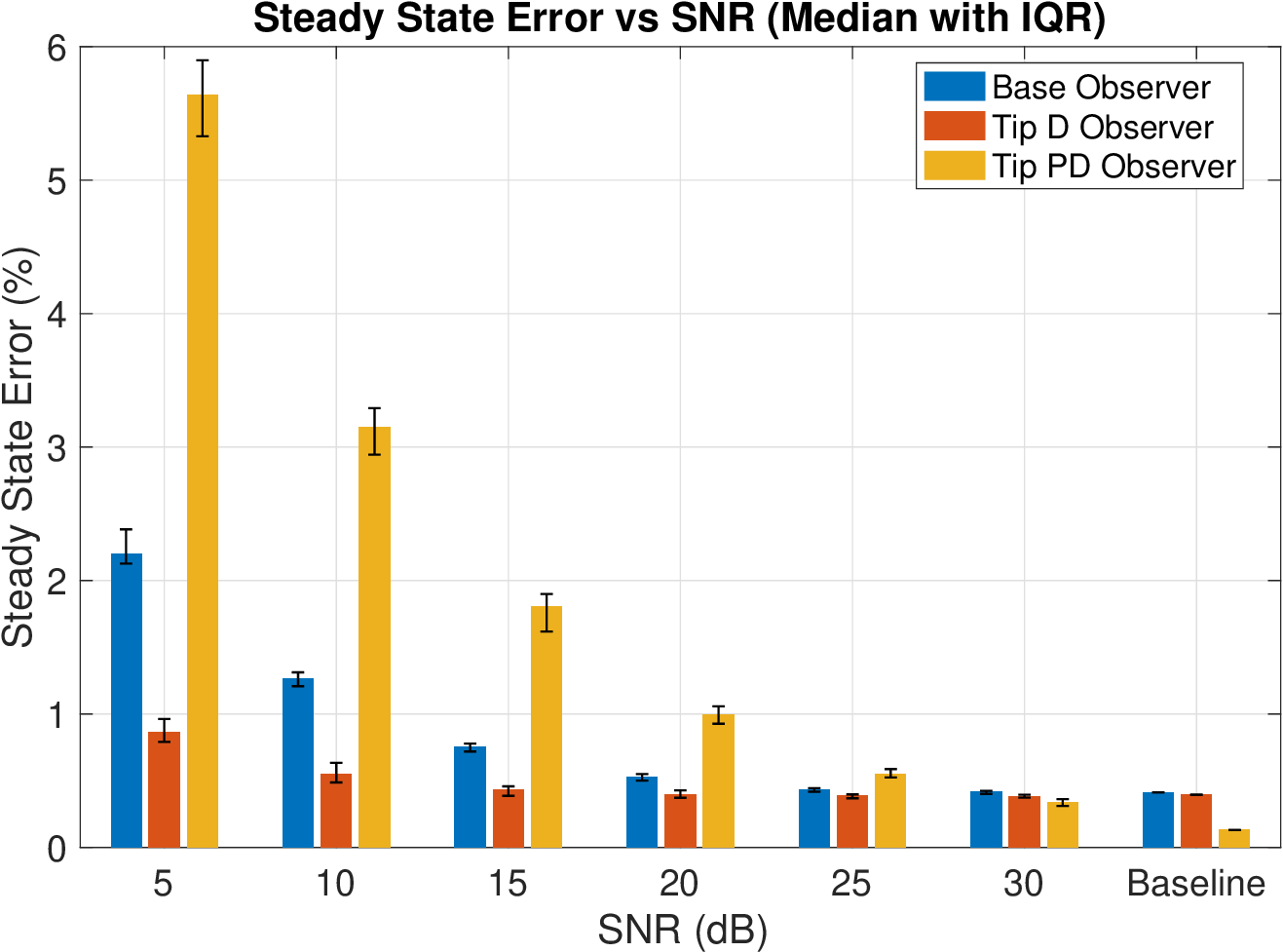}
    \end{subfigure}

    \caption{Error trajectories of the three observers. For each SNR level, all trajectories (32 Monte Carlo runs) are plotted in the same color, resulting in overlapping curves. In the steady-state error plot, each bar represents the median steady-state error (\%) across Monte Carlo runs for a given observer and condition, while the error bars indicate the interquartile range (IQR, 25th--75th percentile).}
    \label{fig:SNR}
\end{figure*}

\begin{figure*}[htbp]
    \centering
    \begin{subfigure}[b]{0.49\textwidth}
        \centering
        \includegraphics[width=\textwidth]{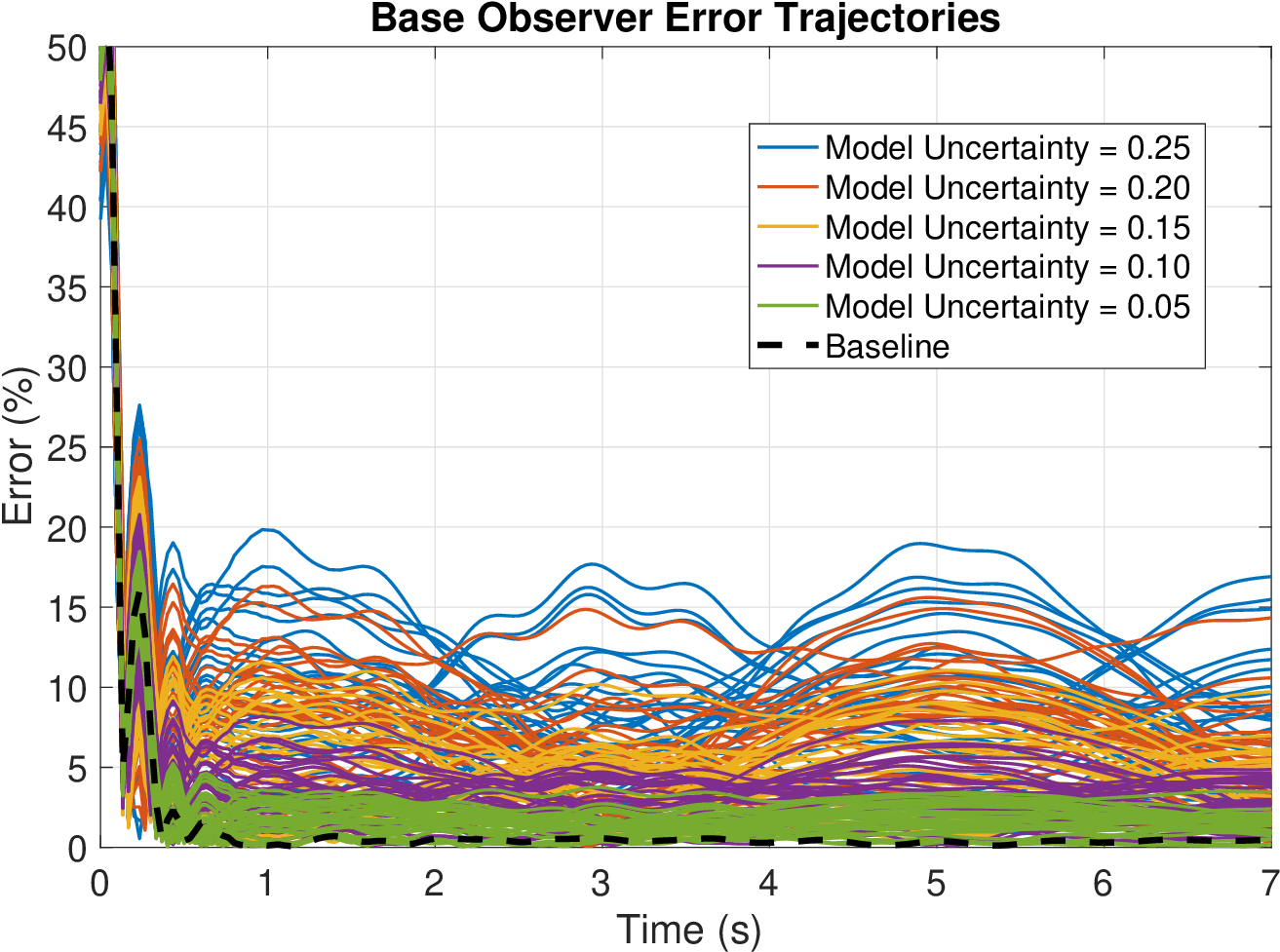}
    \end{subfigure}
    \hfill
    \begin{subfigure}[b]{0.49\textwidth}
        \centering
        \includegraphics[width=\textwidth]{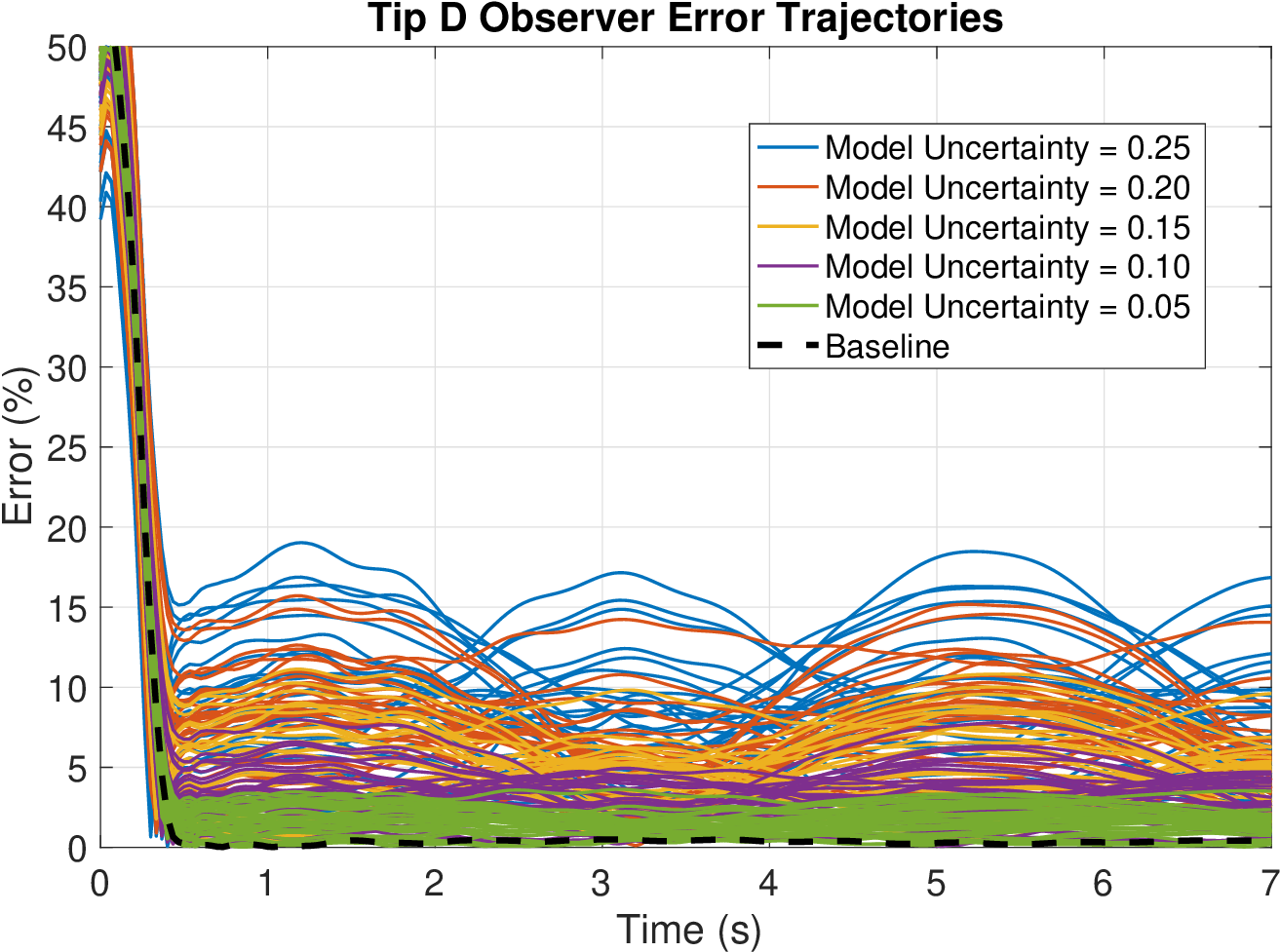}
    \end{subfigure}

    \begin{subfigure}[b]{0.49\textwidth}
        \centering
        \includegraphics[width=\textwidth]{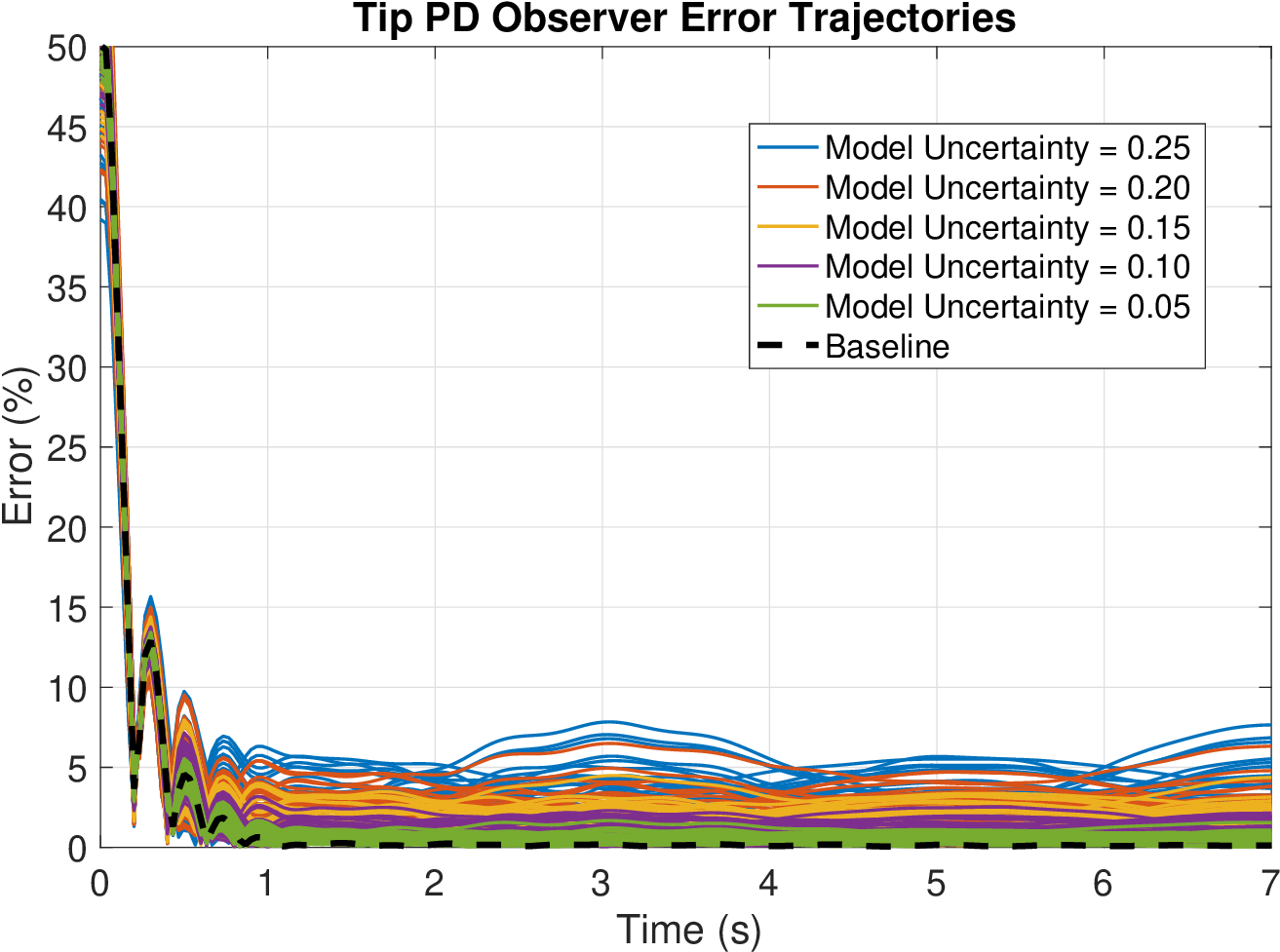}
    \end{subfigure}
    \hfill
    \begin{subfigure}[b]{0.49\textwidth}
        \centering
        \includegraphics[width=\textwidth]{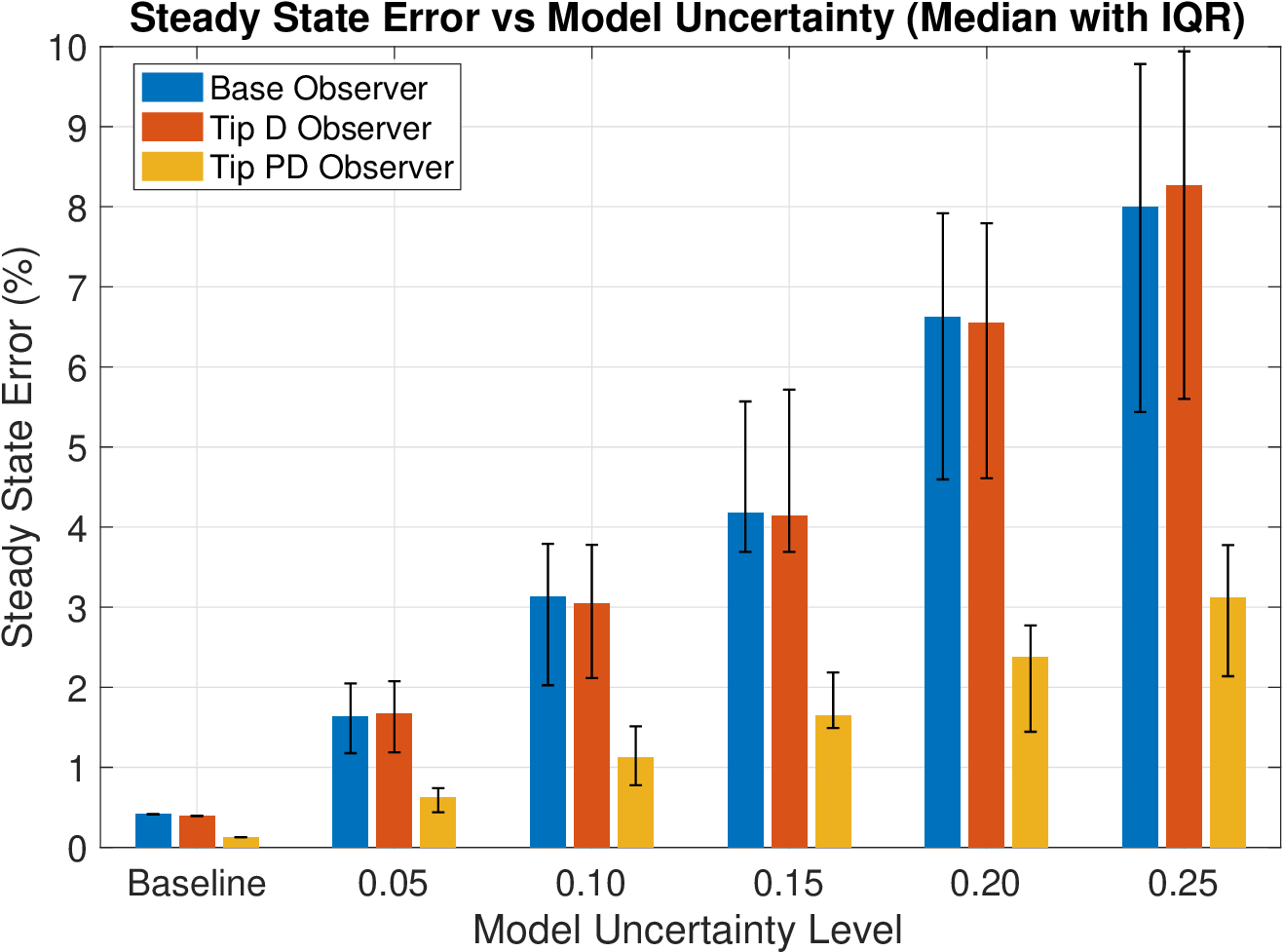}
    \end{subfigure}

    \caption{Error trajectories of the three observers. For each model uncertainty level, all trajectories (32 Monte Carlo runs) are plotted in the same color, resulting in overlapping curves. In the steady-state error plot, each bar represents the median steady-state error (\%) across Monte Carlo runs for a given observer and condition, while the error bars indicate the interquartile range (IQR, 25th--75th percentile).}
    \label{fig:model uncertainty}
\end{figure*}

\section{Monte Carlo Robustness Study}

In this section, we conduct a Monte Carlo study to evaluate the robustness of the boundary observers with respect to measurement noise and model uncertainty separately. The same robot model as in the experimental validation section is used.
The tendon tensions are defined as
\begin{align*}
    \tau_1(t) &= 10\sin\left(\frac{2\pi}{T}t\right), \\
    \tau_2(t) &=
    \begin{cases}
        0, & t < T_s, \\
        10\sin\left(\frac{2\pi}{T}(t - T_s)\right), & t \geq T_s,
    \end{cases}
\end{align*}
where $T = 8$~s is the period and $T_s = 2$~s is a phase shift. This actuation induces a spatial spinning motion of the robot. A viscoelastic damping coefficient of 0.01 was applied when simulating the ground-truth robot motion; otherwise, conservation of total energy would cause sustained oscillations. The damping coefficient was assumed to be unknown, so the observers employed the linear constitutive law \eqref{eq:linear constitutive law}.
The observer gains were set as
\begin{align*}
    \Gamma_0 = \Gamma_0^*, \quad
    \Gamma_1 = \Gamma_1^*, \quad
    \Gamma_{\mathrm{D}} = \Gamma_1^*, \quad
    \Gamma_{\mathrm{P}} = 10\Gamma_1^*.
\end{align*}

Each simulation lasted 8 seconds, and the observers were initialized at $t = 1$~s when the robot was already in motion, resulting in an initial maximum position error exceeding $50\%$ of the robot length.
We define the \emph{error trajectory} as the maximum position error (expressed as a percentage of the robot length) along the arc length. The \emph{steady-state error} is defined as the median error over the final 5 seconds.

\subsection{Robustness to Measurement Noise}

Measurement noise was injected at different signal-to-noise ratios (SNR), defined as
\[
\mathrm{SNR} \in \{5,\,10,\,15,\,20,\,25,\,30\}~\text{dB},
\]
corresponding to strong noise (5–10~dB), moderate noise (15–20~dB), and high-quality measurements (25–30~dB). For each SNR level and each observer, 32 Monte Carlo simulations were performed.

\medskip
\noindent\textbf{Results.}
The results are shown in Fig.~\ref{fig:SNR}, where we plot the error trajectories for all Monte Carlo runs, along with steady-state error statistics.
We observe that measurement noise has little effect on the steady-state error of the base and tip D observers, even at low SNR (5~dB). In contrast, the tip PD observer is more sensitive to noise, and its steady-state error increases as SNR decreases. Notably, the settling time is largely independent of SNR for all observers, indicating that noise has minimal impact on convergence speed.

This difference can be explained by how each feedback signal enters the boundary correction. Boundary wrench and velocity form a power-conjugate pair, and their inner product defines energy exchange at the boundary. When correction is applied through this channel in a dissipative manner, measurement noise enters through an energy-dissipating pathway, resulting only in bounded fluctuations without shifting the equilibrium.
In contrast, pose-based injection introduces a stiffness-like correction rather than acting through a power-conjugate variable. Measurement noise therefore perturbs the boundary equilibrium directly, leading to a noise-dependent steady-state error.

\subsection{Robustness to Model Uncertainty}

Model uncertainty is introduced by perturbing the nominal parameters $M_0$ and $K_0$ as
\begin{align*}
    M &= M_0 \odot \left(1 + \varepsilon\,\Delta_M\right), \\
    K &= K_0 \odot \left(1 + \varepsilon\,\Delta_K\right),
\end{align*}
where $\Delta_M,\Delta_K \in [-1,1]$ are random vectors and
\[
\varepsilon \in \{0.05,\,0.10,\,0.15,\,0.20,\,0.25\}
\]
denotes the model uncertainty level.

The perturbation is applied uniformly along the arc length. This produces larger errors than spatially independent perturbations, since the latter tend to average out during integration. The robot is simulated using perturbed parameters, while the observers use the nominal model.
For each uncertainty level and observer, 32 Monte Carlo simulations were performed.

\medskip
\noindent\textbf{Results.}
The results are shown in Fig.~\ref{fig:model uncertainty}. The steady-state errors of the base and tip D observers increase approximately linearly with model uncertainty and exhibit similar sensitivity. The tip PD observer also shows increasing error but at a slower rate, indicating improved robustness.

This behavior can again be understood by examining the injection mechanisms. Purely dissipative injection (velocity or wrench) preserves passivity but does not define an equilibrium configuration. Under model mismatch, the observer may converge to a biased state.
In contrast, the tip PD observer introduces a stiffness-like boundary constraint. Analogous to a PD controller, the velocity term dissipates dynamic error, while the pose term reduces steady-state bias. This anchors the observer to the measured configuration, reducing drift and improving robustness.

These results reveal a fundamental trade-off: pose feedback improves robustness to model uncertainty but increases sensitivity to measurement noise. Balancing this trade-off requires a rigorous theoretical characterization of the joint effect of noise and model uncertainty, which remains an important direction for future work.

\section{Conclusion and Discussion}
\label{sec:conclusion}

In this work, we presented a novel boundary observer algorithm for state estimation of soft continuum robots, relying only on internal wrench measurements at the robot’s base. This design leverages the mathematical duality between the tip velocity twist and the base internal wrench, and eliminates the need for external motion capture systems. Their combination yields a general formulation of boundary observers. We demonstrated that, in the Kirchhoff rod case, the angular states fully determine the linear states, and vice versa in the planar Kirchhoff case. This inherent redundancy can be exploited to relax sensing requirements in practical implementations. Furthermore, we analyzed the effect of observer gains using Lyapunov-based methods and revealed a non-monotonic ``increase-then-decrease'' trend in the convergence rate when increasing gain magnitude, which is useful for systematic search for optimal gains. 

Extensive simulation and experimental studies on a tendon-driven continuum robot validated the convergence behavior of the proposed observer variants and confirmed the theoretical predictions regarding gain tuning. The observers successfully captured high-frequency oscillations of the robot even with unknown initial conditions. This is a scenario where static estimation algorithms typically struggle. Moreover, with a proportional correction term, the observer remained convergent even when the robot was subjected to completely unknown, time-varying external forces. This robustness is particularly valuable for manipulation tasks involving objects with unknown mass or dynamics.

\medskip
\noindent\textbf{Limitations and Future Work.}
Despite the promising results, several limitations warrant further investigation. First, while the theoretical analysis of optimal observer gains and fastest convergence is quantitatively accurate for a balanced Cosserat rod, it is only qualitatively correct for Kirchhoff rods. A Lyapunov analysis specifically tailored to the Kirchhoff rod model could potentially yield more precise theoretical predictions for optimal gain tuning.

Second, as a model-based approach, the observer relies on a sufficiently accurate system model. The Monte Carlo robustness study further demonstrates performance degradation in the presence of model uncertainty. A natural direction for future work is system identification, where Cosserat rod parameters, including actuator model parameters, are estimated from experimental data. This is particularly important for soft continuum robots that do not possess a well-defined central backbone.

Third, the Monte Carlo study reveals a fundamental trade-off: pose feedback improves robustness to model uncertainty but increases sensitivity to measurement noise. Balancing this trade-off requires a rigorous theoretical characterization of the combined effects of noise and model uncertainty, which remains an open problem.

Finally, observer performance degrades when multiple unknown external forces act simultaneously at different locations along the robot. To the best of our knowledge, state estimation in the presence of dynamic motion and multiple unknown external forces remains an open problem. Extending the proposed boundary observers to such scenarios, potentially with more measurements, would be an important research direction.


\section*{Appendix: Notation in $SO(3)$ and $SE(3)$}
\label{sec:notation}

Denote by $SO(3)$ the special orthogonal group (the group of rotation matrices) and by $so(3)$ its associated Lie algebra.
Denote by $SE(3)=SO(3)\times\mathbb{R}^3$ the special Euclidean group (the group of homogeneous transformation matrices) and by $se(3)$ its associated Lie algebra.
A hat $\wedge$ in the superscript of a vector $\eta$ defines a matrix $\eta\cross$ whose definition depends on the dimension of $\eta$.
Specifically, if $\eta\in\mathbb{R}^3$, then $\eta\cross\in so(3)$ is such that $\eta\cross\xi=\eta\times\xi$ for any $\xi\in\mathbb{R}^3$ where $\times$ is the cross product.
In this case, $\eta\cross$ turns out to be a skew-symmetric matrix in $\mathbb{R}^{3\times3}$.
If $\eta=[w^\top,v^\top]^\top\in\mathbb{R}^6$ with $w,v\in\mathbb{R}^3$, then $\eta\cross\in se(3)$ is defined by
\begin{align*}
    \eta\cross=
    \begin{bmatrix}
        w\cross & v \\
        0_{1\times3} & 0
    \end{bmatrix}.
\end{align*}
The adjoint operator $\ad$ of $\eta=[w^\top~v^\top]^\top\in\mathbb{R}^6$ with $w,v\in\mathbb{R}^3$ is defined by
\begin{align*}
    \ad_\eta=
    \begin{bmatrix}
        w\cross & 0_{3\times3} \\
        v\cross & w\cross
    \end{bmatrix}.
\end{align*}
Let the superscript $\vee$ be the inverse operator of $\wedge$, i.e., $(\eta\cross)^\vee=\eta$.
The adjoint representation of any element $g=(R,p)\in SE(3)$ with $R\in SO(3)$ and $p\in\mathbb{R}^3$ is defined by
\begin{align*}
    \Ad_g=
    \begin{bmatrix}
        R & 0_{3\times3} \\
        p\cross R & R
    \end{bmatrix}\in\mathbb{R}^{6\times6}.
\end{align*}

\section*{Acknowledgments}
The authors thank Chloe Pogue, Puspita Triana Dewi, and Chengnan (Jimmy) Shentu in the Continuum Robotics Laboratory for their assistance with the experiments. The authors are especially grateful to Yuting Zhang for identifying and helping correct a crucial error in the experimental implementation.

\section*{Funding}
The authors acknowledge the funding support of the Natural Sciences and Engineering Research Council of Canada (NSERC, RGPIN-2025-04343), the Canada Foundation for Innovation (CFI), and the Ontario Research Fund (ORF).

\section*{Declaration of conflicting interests}
The authors declared no potential conflicts of interest with respect to the research, authorship, and/or publication of this article.

\bibliographystyle{SageH}
\bibliography{references}

@inproceedings{robinson1999ContinuumRobotsStatea,
  title = {Continuum Robots - a State of the Art},
  booktitle = {{IEEE International Conference} on {{Robotics}} and {{Automation}} },
  author = {Robinson, G. and Davies, J.B.C.},
  year = {1999},
  volume = {4},
  pages = {2849--2854}
}

@article{burgner2015continuum,
  title={Continuum robots for medical applications: A survey},
  author={Burgner-Kahrs, Jessica and Rucker, D Caleb and Choset, Howie},
  journal={IEEE Transactions on Robotics},
  volume={31},
  number={6},
  pages={1261--1280},
  year={2015}
}

@article{russo2023ContinuumRobotsOverview,
  title = {Continuum {{Robots}}: {{An Overview}}},
  shorttitle = {Continuum {{Robots}}},
  author = {Russo, Matteo and Sadati, Seyed Mohammad Hadi and Dong, Xin and Mohammad, Abdelkhalick and Walker, Ian D. and Bergeles, Christos and Xu, Kai and Axinte, Dragos A.},
  year = {2023},
  journal = {Advanced Intelligent Systems},
  volume = {5},
  number = {5},
  pages = {2200367}
}

@article{song2015electromagnetic,
  title={Electromagnetic positioning for tip tracking and shape sensing of flexible robots},
  author={Song, Shuang and Li, Zheng and Yu, Haoyong and Ren, Hongliang},
  journal={IEEE Sensors Journal},
  volume={15},
  number={8},
  pages={4565--4575},
  year={2015}
}

@inproceedings{bezawada2022shape,
  title={Shape Estimation of Soft Manipulators using Piecewise Continuous Pythagorean-Hodograph Curves},
  author={Bezawada, Harish and Woods, Cole and Vikas, Vishesh},
  booktitle={American Control Conference},
  pages={2905--2910},
  year={2022}
}

@article{sadati2020stiffness,
  title={Stiffness imaging with a continuum appendage: Real-time shape and tip force estimation from base load readings},
  author={Sadati, SM Hadi and Shiva, Ali and Herzig, Nicolas and Rucker, Caleb D and Hauser, Helmut and Walker, Ian D and Bergeles, Christos and Althoefer, Kaspar and Nanayakkara, Thrishantha},
  journal={IEEE Robotics and Automation Letters},
  volume={5},
  number={2},
  pages={2824--2831},
  year={2020}
}

@article{feliu2025actuation,
  title = {Actuation Reading Insights: Estimating Shape and Forces in Tendon-Driven Slender Soft Robots},
  author = {Feliu-Talegon, Daniel and Alkayas, Abdulaziz Y. and Adamu, Yusuf Abdullahi and Mathew, Anup Teejo and Renda, Federico},
  year = {2025},
  journal = {IEEE/ASME Transactions on Mechatronics},
  volume = {30},
  number = {6},
  pages = {7878--7888}
}

@article{anderson2017continuum,
  title={Continuum reconfigurable parallel robots for surgery: Shape sensing and state estimation with uncertainty},
  author={Anderson, Patrick L and Mahoney, Arthur W and Webster, Robert James},
  journal={IEEE Robotics and Automation Letters},
  volume={2},
  number={3},
  pages={1617--1624},
  year={2017}
}

@article{lilge2022continuum,
  title={Continuum robot state estimation using Gaussian process regression on SE(3)},
  author={Lilge, Sven and Barfoot, Timothy D and Burgner-Kahrs, Jessica},
  journal={The International Journal of Robotics Research},
  volume={41},
  number={13-14},
  pages={1099--1120},
  year={2022}
}

@article{ferguson2024unified,
  title={Unified shape and external load state estimation for continuum robots},
  author={Ferguson, James M and Rucker, D Caleb and Webster, Robert J},
  journal={IEEE Transactions on Robotics},
  volume={40},
  pages={1813--1827},
  year={2024}
}

@inproceedings{loo2019h,
  title={H-infinity based extended kalman filter for state estimation in highly non-linear soft robotic system},
  author={Loo, Junn Yong and Tan, Chee Pin and Nurzaman, Surya Girinatha},
  booktitle={American Control Conference},
  pages={5154--5160},
  year={2019}
}

@inproceedings{rucker2022task,
  title={Task-Space Control of Continuum Robots using Underactuated Discrete Rod Models},
  author={Rucker, Caleb and Barth, Eric J and Gaston, Joshua and Gallentine, James C},
  booktitle={IEEE/RSJ International Conference on Intelligent Robots and Systems},
  pages={10967--10974},
  year={2022}
}

@article{feliu2024dynamic,
  title={Dynamic shape estimation of tendon-driven soft manipulators via actuation readings},
  author={Feliu-Talegon, Daniel and Mathew, Anup Teejo and Alkayas, Abdulaziz Y and Adamu, Yusuf Abdullahi and Renda, Federico},
  journal={IEEE Robotics and Automation Letters},
  volume={10},
  number={1},
  pages={780--787},
  year={2025}
}

@article{zheng2024full,
  author={Zheng, Tongjia and Han, Qing and Lin, Hai},
  journal={IEEE Transactions on Automatic Control}, 
  title={Full State Estimation of Continuum Robots From Tip Velocities: A Cosserat-Theoretic Boundary Observer}, 
  year={2025},
  volume={70},
  number={5},
  pages={2859-2872}
}

@inproceedings{zheng2024estimating,
  title={Estimating Infinite-Dimensional Continuum Robot States From the Tip},
  author={Zheng, Tongjia and McFarland, Ciera and Coad, Margaret and Lin, Hai},
  booktitle={IEEE International Conference on Soft Robotics},
  pages={572--578},
  year={2024}
}

@article{simo1988dynamics,
  title={On the dynamics in space of rods undergoing large motions—a geometrically exact approach},
  author={Simo, Juan Carlos and Vu-Quoc, Loc},
  journal={Computer methods in applied mechanics and engineering},
  volume={66},
  number={2},
  pages={125--161},
  year={1988},
  publisher={Elsevier}
}

@article{rucker2011statics,
  title={Statics and dynamics of continuum robots with general tendon routing and external loading},
  author={Rucker, D Caleb and Webster III, Robert J},
  journal={IEEE Transactions on Robotics},
  volume={27},
  number={6},
  pages={1033--1044},
  year={2011}
}

@article{renda2014dynamic,
  title={Dynamic model of a multibending soft robot arm driven by cables},
  author={Renda, Federico and Giorelli, Michele and Calisti, Marcello and Cianchetti, Matteo and Laschi, Cecilia},
  journal={IEEE Transactions on Robotics},
  volume={30},
  number={5},
  pages={1109--1122},
  year={2014}
}

@article{della2020model,
  title={Model-based dynamic feedback control of a planar soft robot: trajectory tracking and interaction with the environment},
  author={Della Santina, Cosimo and Katzschmann, Robert K and Bicchi, Antonio and Rus, Daniela},
  journal={The International Journal of Robotics Research},
  volume={39},
  number={4},
  pages={490--513},
  year={2020}
}

@article{renda2018discrete,
  title={Discrete cosserat approach for multisection soft manipulator dynamics},
  author={Renda, Federico and Boyer, Fr{\'e}d{\'e}ric and Dias, Jorge and Seneviratne, Lakmal},
  journal={IEEE Transactions on Robotics},
  volume={34},
  number={6},
  pages={1518--1533},
  year={2018}
}

@article{boyer2020dynamics,
  title={Dynamics of continuum and soft robots: A strain parameterization based approach},
  author={Boyer, Frederic and Lebastard, Vincent and Candelier, Fabien and Renda, Federico},
  journal={IEEE Transactions on Robotics},
  volume={37},
  number={3},
  pages={847--863},
  year={2020}
}

@book{murray2017mathematical,
  title={A mathematical introduction to robotic manipulation},
  author={Murray, Richard M and Li, Zexiang and Sastry, S Shankar},
  year={2017},
  publisher={CRC press}
}

@article{renda2020geometric,
  title={A geometric variable-strain approach for static modeling of soft manipulators with tendon and fluidic actuation},
  author={Renda, Federico and Armanini, Costanza and Lebastard, Vincent and Candelier, Fabien and Boyer, Frederic},
  journal={IEEE Robotics and Automation Letters},
  volume={5},
  number={3},
  pages={4006--4013},
  year={2020}
}

@article{boyer2022statics,
  title={Statics and Dynamics of Continuum Robots Based on Cosserat Rods and Optimal Control Theories},
  author={Boyer, Fr{\'e}d{\'e}ric and Lebastard, Vincent and Candelier, Fabien and Renda, Federico and Alamir, Mazen},
  journal={IEEE Transactions on Robotics},
  volume={39},
  number={2},
  pages={1544--1562},
  year={2023}
}

@article{rao2021model,
  title={How to model tendon-driven continuum robots and benchmark modelling performance},
  author={Rao, Priyanka and Peyron, Quentin and Lilge, Sven and Burgner-Kahrs, Jessica},
  journal={Frontiers in Robotics and AI},
  volume={7},
  pages={630245},
  year={2021}
}

@article{till2019real,
  title={Real-time dynamics of soft and continuum robots based on Cosserat rod models},
  author={Till, John and Aloi, Vincent and Rucker, Caleb},
  journal={The International Journal of Robotics Research},
  volume={38},
  number={6},
  pages={723--746},
  year={2019}
}

@article{mathew2022sorosim,
  title={SoRoSim: A MATLAB Toolbox for Hybrid Rigid-Soft Robots Based on the Geometric Variable-Strain Approach},
  author={Mathew, Anup Teejo and Hmida, Ikhlas Mohamed Ben and Armanini, Costanza and Boyer, Frederic and Renda, Federico},
  journal={IEEE Robotics \& Automation Magazine},
  year={2023},
  volume={30},
  number={3},
  pages={106-122}
}

@book{choset2005PrinciplesRobotMotion,
  title = {Principles of Robot Motion: Theory, Algorithms, and Implementations},
  shorttitle = {Principles of Robot Motion},
  author = {Choset, Howie and Lynch, Kevin M. and Hutchinson, Seth and Kantor, George A. and Burgard, Wolfram and Kavraki, Lydia E. and Thrun, Sebastian},
  year = {2005},
  address = {Cambridge, MA},
  publisher = {MIT Press}
}

@book{barfoot2017state,
  title={State estimation for robotics},
  author={Barfoot, Timothy D},
  year={2017},
  publisher={Cambridge University Press}
}

@inproceedings{brace2022sensor,
  title={Sensor placement on a cantilever beam using observability gramians},
  author={Brace, Natalie L and Andrews, Nicholas B and Upsal, Jeremy and Morgansen, Kristi A},
  booktitle={IEEE 61st Conference on Decision and Control},
  pages={388--395},
  year={2022}
}

@book{krstic2008boundary,
  title={Boundary control of PDEs: A course on backstepping designs},
  author={Krstic, Miroslav and Smyshlyaev, Andrey},
  volume={16},
  year={2008},
  publisher={Siam}
}

@inproceedings{vazquez2011backstepping,
  title={Backstepping boundary stabilization and state estimation of a {$2\times 2$} linear hyperbolic system},
  author={Vazquez, Rafael and Krstic, Miroslav and Coron, Jean-Michel},
  booktitle={50th IEEE Conference on Decision and Control and European Control Conference},
  pages={4937--4942},
  year={2011},
  organization={IEEE}
}

@article{castillo2013boundary,
  title={Boundary observers for linear and quasi-linear hyperbolic systems with application to flow control},
  author={Castillo, Felipe and Witrant, Emmanuel and Prieur, Christophe and Dugard, Luc},
  journal={Automatica},
  volume={49},
  number={11},
  pages={3180--3188},
  year={2013},
  publisher={Elsevier}
}

@book{evans1988partial,
  title={Partial Differential Equations},
  author={Evans, Lawrence C},
  edition={2},
  series={Graduate Studies in Mathematics},
  volume={19},
  year={2010},
  publisher={American Mathematical Society}
}

@incollection{lee2003smooth,
  title={Smooth manifolds},
  author={Lee, John M},
  booktitle={Introduction to smooth manifolds},
  pages={1--29},
  year={2003},
  publisher={Springer}
}

@book{bastin2016stability,
  title={Stability and boundary stabilization of 1-d hyperbolic systems},
  author={Bastin, Georges and Coron, Jean-Michel},
  volume={88},
  year={2016},
  publisher={Springer}
}

@article{janabi2021cosserat,
  title={Cosserat Rod-Based Dynamic Modeling of Tendon-Driven Continuum Robots: A Tutorial},
  author={Janabi-Sharifi, Farrokh and Jalali, Amir and Walker, Ian D},
  journal={IEEE Access},
  volume={9},
  pages={68703--68719},
  year={2021}
}

\end{document}